\documentclass[sigconf]{acmart}

\AtBeginDocument{%
  \providecommand\BibTeX{{%
    \normalfont B\kern-0.5em{\scshape i\kern-0.25em b}\kern-0.8em\TeX}}}

\copyrightyear{2026}
\acmYear{2026}
\setcopyright{cc}
\setcctype{by}
\acmConference[KDD '26]{Proceedings of the 32nd ACM SIGKDD Conference on Knowledge Discovery and Data Mining V.2}{August 09--13, 2026}{Jeju Island, Republic of Korea}
\acmBooktitle{Proceedings of the 32nd ACM SIGKDD Conference on Knowledge Discovery and Data Mining V.2 (KDD '26), August 09--13, 2026, Jeju Island, Republic of Korea}
\acmDOI{10.1145/3770855.3818437}
\acmISBN{979-8-4007-2259-2/2026/08}

\usepackage{optidef}
\usepackage{enumerate}
\usepackage{subcaption}
\usepackage{enumitem}
\newcommand{\blue}[1]{{#1}}

\def\method{\text{JourneyFormer}}

\begin{document}

\title{\method{}: Encoding Airbnb Guest Journey with Sequence Modeling}



\author{Daochen Zha}
\affiliation{%
  \institution{Airbnb}
  \city{San Francisco}
  \country{USA}}
\email{daochen.zha@airbnb.com}

\author{Chun How Tan}
\affiliation{%
  \institution{Airbnb}
  \city{San Francisco}
  \country{USA}}
\email{chunhow.tan@airbnb.com}

\author{Xin Liu}
\affiliation{%
  \institution{Airbnb}
  \city{San Francisco}
  \country{USA}}
\email{xin.liu@airbnb.com}

\author{Bin Xu}
\affiliation{%
  \institution{Airbnb}
  \city{San Francisco}
  \country{USA}}
\email{b.xu@airbnb.com}

\author{Han Zhao}
\affiliation{%
  \institution{Airbnb}
  \city{San Francisco}
  \country{USA}}
\email{han.zhao@airbnb.com}

\author{Xiaowei Liu}
\affiliation{%
  \institution{Airbnb}
  \city{San Francisco}
  \country{USA}}
\email{xiaowei.liu@airbnb.com}

\author{Tracy Yu}
\affiliation{%
  \institution{Airbnb}
  \city{San Francisco}
  \country{USA}}
\email{tracy.yu@airbnb.com}

\author{Hui Gao}
\affiliation{%
  \institution{Airbnb}
  \city{San Francisco}
  \country{USA}}
\email{hui.gao@airbnb.com}

\author{Huiji Gao}
\affiliation{%
  \institution{Airbnb}
  \city{San Francisco}
  \country{USA}}
\email{huiji.gao@airbnb.com}

\author{Liwei He}
\affiliation{%
  \institution{Airbnb}
  \city{San Francisco}
  \country{USA}}
\email{liwei.he@airbnb.com}

\author{Stephanie Moyerman}
\affiliation{%
 \institution{Airbnb}
 \city{San Francisco}
 \country{USA}}
\email{stephanie.moyerman@airbnb.com}

\author{Sanjeev Katariya}
\affiliation{%
 \institution{Airbnb}
 \city{San Francisco}
 \country{USA}}
\email{sanjeev.katariya@airbnb.com}

\renewcommand{\shortauthors}{Daochen Zha, et al.}

\begin{abstract}
Sequence modeling has become increasingly popular in recommendation and ranking algorithms, owing to its capacity to model users’ historical behaviors and infer user intentions. Despite its theoretical simplicity, the practical deployment of a sequence model in production is non-trivial due to complexity of the sequence and sparse labels. For example, in Airbnb, guest sequences are often long, exploratory and complex, and we focus on booking labels, which are sparse. As such, we are often required to make various design decisions regarding data and modeling to strike a balance between effectiveness and scalability. This work delved into these production challenges and deployed \method{}, a sequence modeling solution for search ranking at Airbnb. We detail crucial design considerations, covering aspects such as guest event selection, ID embeddings, model architecture, and label attribution. Additionally, we describe several tailored strategies to accelerate model training and inference. \method{} has been successfully deployed within Airbnb's production, where its effectiveness and impact have been evidenced not only by improved offline ranking metrics but also by significant gains in key business metrics through online A/B testing across 2 production surfaces.

\end{abstract}






\maketitle

\section{Introduction} \label{sec:1}

On two-sided e-commerce platforms like Airbnb, search ranking plays a vital role. Guests typically begin their journey on Airbnb with a search, then browse and interact with the ranked listings. A smart search ranking algorithm can help guests find suitable stays and make confident booking decisions, a triple win for guests, hosts, and the platform: guests find what they are looking for, hosts receive more bookings, and the overall business on Airbnb grows.

\begin{figure*}[t]
  \centering
  \includegraphics[width=\textwidth]{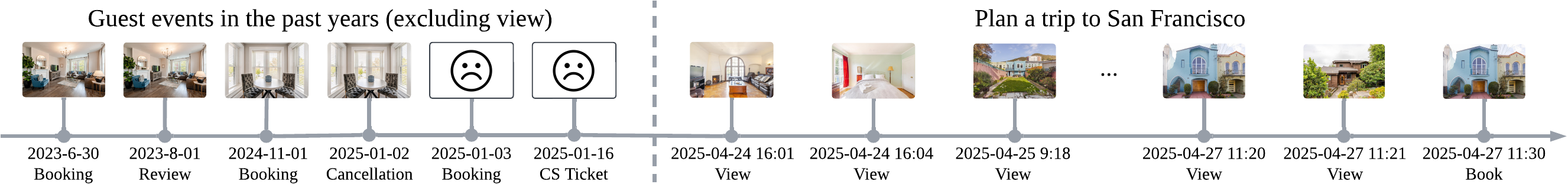}
  \caption{An example of a guest journey—typically long, exploratory, and complex. Note that this journey is purely hypothetical for illustration purpose and is not from any real guest data. The sad face refers to a listing that delivered a poor guest experience.}
  \label{fig:toy}
\end{figure*}

Booking decisions are complicated as guests often invest considerable time browsing and comparing listings. Figure~\ref{fig:toy} shows an example of a guest journey, which is typically long, exploratory, and complex. Beyond viewing  (clicking to open the listing page) and booking events, guests may cancel trips (e.g., due to a change of plans), leave reviews (positive or negative), or contact Customer Service (CS). These events naturally form a time-stamped sequence.

Sequence modeling has recently gained traction in recommendation and ranking algorithms~\cite{sun2019bert4rec,kang2018self,wu2020sse,li2020time,chen2022intent,fan2022sequential,du2023frequency,zhang2023adaptive,zhou2023attention,Zhai2024ActionsSL,pancha2022pinnerformer}. The core idea of these methods is to use sequence models like Transformer~\cite{vaswani2017attention} to encode user sequence into compact representations. These representations are then used to enhance recommendation and ranking performance. Inspired by the impressive results in prior research, we aim to encode Airbnb’s guest journeys with sequence modeling to better infer guest intent and surface the right listings in search ranking.

However, despite its theoretical simplicity, the practical deployment of a sequence model in production is non-trivial due to complexity of the sequence and sparse labels. For example, in Airbnb search ranking, there are several challenges:
\begin{itemize}[leftmargin=*]
    \item Long and exploratory nature of the guest journey - Many guests spend significant time exploring Airbnb listings across different destinations and dates by interacting with the search ranking, often resulting in a very long guest sequences. Some guest events may reflect an intent to find a place to stay, while others could be simply exploratory without a clear goal. Additionally, the user might exhibit back-and-forth behavior when viewing the same listing, which is often uncommon on social media platforms. Extracting meaningful representations from these long, noisy sequences is challenging.
    \item Complexity of guest events - Guest events come in many forms: some indicate preference (e.g., bookings), while others are ambiguous (e.g., reviews, views, or CS tickets). Their frequency also varies widely—views, for instance, are far more common than other events. A sequence model must effectively capture this complexity to model guest behavior well.
    \item Focusing on the sparse booking labels rather than engagement - Unlike social media platforms~\cite{pancha2022pinnerformer,Xia2023TransActTR} that often focus on engagement, e-commerce platforms like Airbnb prioritize booking conversion. As a result, engagement signals can be noisy. For instance, a listing view might reflect genuine booking interest—or simply as a comparison to other listings, or even random browsing. Moreover, guests rarely book the same or sometimes even similar listings twice, so suggesting similar options is often not effective. We need a truly generalizable model to infer guests’ underlying needs.
\end{itemize}


As a result, practically deploying a sequence model in a production environment like in Airbnb typically faces a wide range of obstacles and tradeoffs—from feature formulation and label definition to training and inference scalability. Our work aims to bridge this gap and facilitate the real-world adoption of sequence models in industry-scale search ranking platforms.

In this work, we present \method{}, a sequence modeling solution deployed in Airbnb’s search ranking. While most prior work is research-oriented and focuses on improving prediction accuracy, ours addresses the production challenges of deploying sequence models—which often require careful trade-offs. To keep our contribution focused, we adopt a simple modeling architecture and center our discussions on key design decisions encountered in production, such as guest event selection, ID embeddings, model architecture, label attribution, and strategies to accelerate training and inference. In summary, our main contributions are:

\begin{itemize}[leftmargin=*]
    \item We discuss sequence model beyond model architecture itself, covering key production considerations—from selecting and capping guest events, to feature encoding, label attribution, and strategies for efficient training and model serving.
    \item We successfully deployed \method{}, a sequence model in Airbnb’s search ranking. Unlike our previous ranking models that relied on hand-crafted features, \method{} uses a Transformer to encode guest journeys, resulting in more expressive and meaningful guest representations.
    \item \method{} not only achieves substantial improvements in offline metrics over a strong baseline, but also delivers significant gains in key business metrics—such as uncanceled bookers and nights booked—through online A/B testing across 2 production surfaces on search ranking and promotional emails. Moreover, we conduct extensive analysis to understand \method{} under alternative design choices, the impact of ablated training acceleration strategies, performance across sequence lengths, and sensitivity to distribution shifts.
    \item Furthermore, we discuss several seemingly promising directions that we explored but ultimately did not adopt. We hope our lessons and findings can benefit applied researchers and help the broader research community understand the complexities faced in industrial search ranking applications.
\end{itemize}

\section{Preliminaries} \label{sec:2}

\subsection{How is Guest Journey Encoded at Airbnb Search Ranking prior to \method{}?}
\label{sec:2:1}

The idea of leveraging the guest journey to enhance search ranking is not new. Prior to \method{}, Airbnb's ranking algorithms relied on hand-crafted features to encapsulate the guest journey~\cite{tan2023optimizing,tang2024multi,haldar2019applying}. These features were typically devised by aggregating various guest statistics across different time frames. For instance, features such as the total number of bookings and the average price of bookings over a specified period were common. Additionally, some guest-listing crossed features, like the number of times a guest viewed a specific listing within a specific period, were also used. The hand-crafted features offered conveniences in production: the feature aggregation can be straightforwardly implemented through either batched or real-time data pipelines; for inference, the models often employed neural network architectures like MLP, which are relatively cost-efficient and easy to support.

However, our team is facing scalability challenges with the above hand-crafted feature engineering approach. Specifically, as machine learning engineers kept adding more and more features into the model, maintaining the data pipelines became increasingly difficult. Furthermore, the hand-crafted nature of these features could result in sub-optimal performance, as the model might struggle to infer complex guest intentions from the aggregated feature values alone. This motivated us to explore a more scalable and sustainable search ranking architecture based on sequence modeling.

\begin{figure*}[t]
  \centering
  \includegraphics[width=\textwidth]{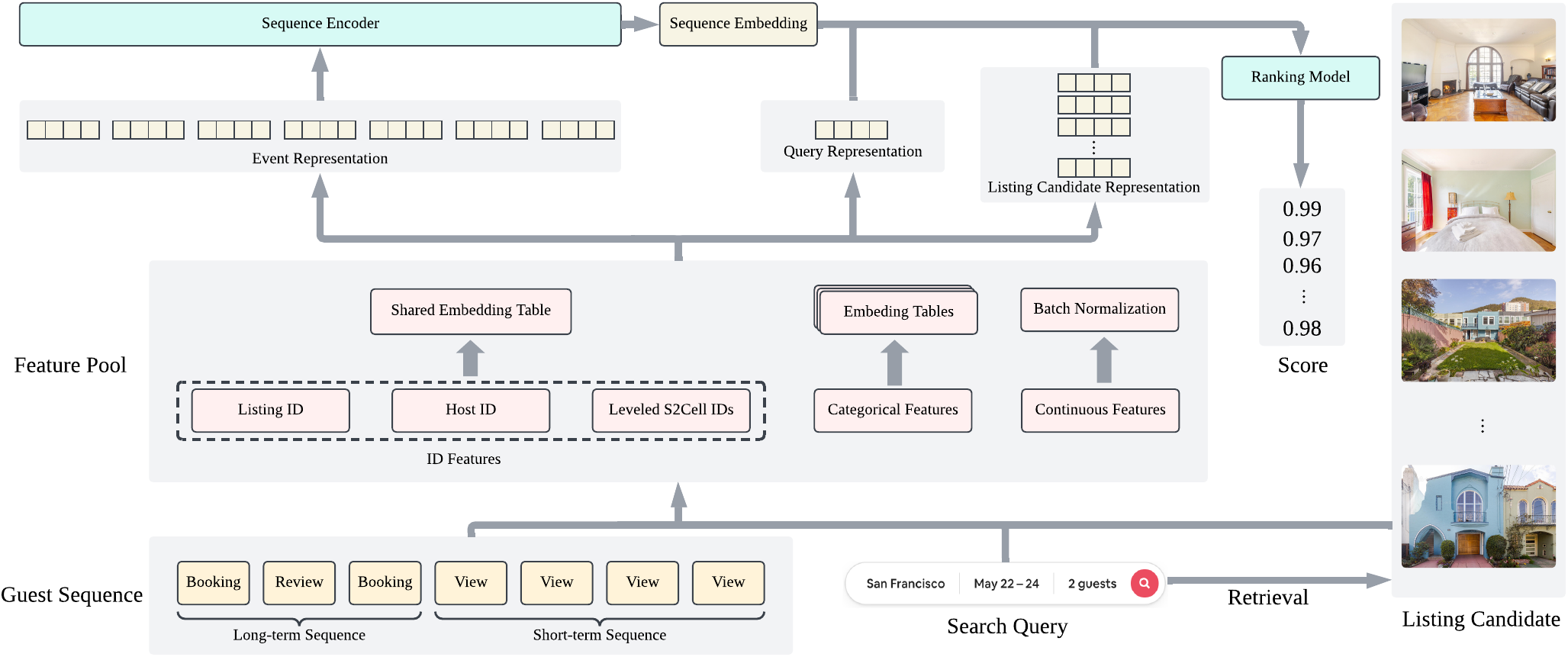}
  \caption{An overview of \method{}. The guest sequence consists of a long-term sequence (events from the past 7 years, excluding frequent events like listing views) and a short-term sequence (frequent events from the past 21 days). The guest sequence, the search query and the retrieved listing candidates use a shared feature pool to construct feature representations. They are then processed by a sequence encoder and a ranking model to generate listing scores.}
  \label{fig:overview}
\end{figure*}

\subsection{Notations \& Problem Statement}

We start by considering a guest $g \in \mathcal{G}$ initiates a search at Airbnb, where $\mathcal{G}$ denotes the set of all Airbnb guests. The search request includes a query, denoted as $q$, encompassing information like destinations, dates, and guest numbers. The search backend then retrieves relevant listings $\mathcal{L}_q$ from Airbnb's listing pool $\mathcal{L}$ ($|\mathcal{L}| > 8M$) using $q$, where $\mathcal{L}_q = \{l_1, l_2, ..., l_n\} \subset \mathcal{L}$, and $n$ varies significantly from query to query, typically falling within the range of tens of thousands. Additionally, we can retrieve a sequence of guest events arranged in chronological order, denoted as $\mathcal{S}_g = \{e_1, e_2, ..., e_t\}$, where each $e_i \in \mathcal{E}$ ($i \in \{1, 2, ..., t\}$). The set $\mathcal{E}$ encompasses all events logged by the platform, such as listing views, booking requests, and reviews. $|\mathcal{E}|$ is quite large, typically, containing hundreds of millions of events on a daily basis. The length of the guest event sequence $|\mathcal{S}_g|$ differs greatly among guests, spanning from 0 (indicating a new guest) to hundreds of thousands.

Rather than using the aggregated features to represent $\mathcal{S}_g$ as discussed in Section~\ref{sec:2:1}, our goal is to learn a sequence embedding $\mathbf{e}_g$ for each guest and use it to improve ranking results. More specifically, given the guest event sequence $\mathcal{S}_g$, query $q$, and retrieved listings $\mathcal{L}_q$, we aim to develop a sequence model that can rank the listings in $\mathcal{L}_q$ to boost conversions; in Airbnb's search ranking context, one of the primary focus is on increasing uncancelled bookings and uncancelled nights booked.

\section{\method{}} \label{sec:3}

Figure~\ref{fig:overview} shows an overview of \method{}. It selects a subset of events to construct long-term and short-term sequences based on the frequency of each event type (Section~\ref{sec:3:1}). Then a shared feature pool is used for guest sequences, search queries, and listing candidates to derive feature representations (Section~\ref{sec:3:2}). For modeling, a sequence encoder is employed to generate sequence embeddings (Section~\ref{sec:3:3}), followed by a ranking model to score listing candidates (Section~\ref{sec:3:4}). Furthermore, we introduce several training strategies that helped boost training throughput by around 4x (Section~\ref{sec:3:5}), as well as the model serving workflow which combines batch inference and real-time inference (Section~\ref{sec:3:6}).

\subsection{Long-term \& Short-term Sequences}
\label{sec:3:1}
We consider the following event types in a guest journey:

\begin{itemize}[leftmargin=*]
    \item View - An opening of a product listing page. Note that, on social media platforms, "views" typically indicate that a user has seen an item in a results list before clicking. On Airbnb, however, a "view" refers to a user clicking into the listing page.
    \item Booking request - A listing reservation request from a guest.
    \item Booking - An acceptance (by host) of a reservation request.
    \item Cancellation - A cancellation of a reservation.
    \item Review - A review for a listing from a guest.
    \item CS ticket - A customer service ticket from a guest or host.
\end{itemize}
Other events such as wishlist creation, map movements, and payment page visits could also be included in the \method{} framework. We will explore them in the future work.

\begin{figure}[t]
  \centering
  \includegraphics[width=0.46\textwidth]{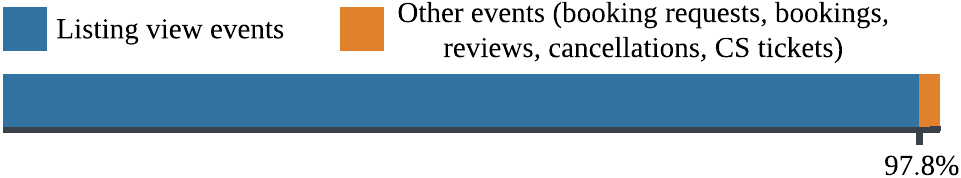}
  \caption{The distribution of event types, with the majority being listing views, while the others only account for 2.2\%.}
  \label{fig:event_distribution}
\end{figure}

\textbf{Event selection.} Modeling the guest sequence $\mathcal{S}_g$ is challenging due to its typically large size, with some certain guests having an extremely large number of events. To reduce the cost, we adopt a \emph{frequency-based selection strategy}. Specifically, for frequent events, we retain only those within a narrow time window; for infrequent events, we use a broader time frame to preserve most of them. To instantiate this strategy, we plot the event type distribution in Figure~\ref{fig:event_distribution}, which shows that the majority is view event. Motivated by this, we construct two guest sequences as follows:

\begin{itemize}[leftmargin=*]
    \item Long-term sequence - This includes the events from the past 7 years, focusing on infrequent events (i.e., excluding view events).
    \item Short-term sequence - This only includes the view events from the guest in the past 21 days.
\end{itemize}

\textbf{Event capping.} To ensure scalability, we limit long-term and short-term sequences to a maximum of 80 and 200 recent events, respectively. The thresholds are set by limiting the top 2\% of guests with the longest sequences and taking into account of the GPU VRAM usage necessary for model training. While raising these thresholds could improve ranking performance, it may impact scalability. We aim to investigate this further in the future. The final guest sequence is obtained by combining the capped long-term and short-term events and then reordering them using time-stamps.

\textbf{Discussion.} In e-commerce platforms like Airbnb, infrequent events can be highly informative. For instance, a booking request could clearly indicate guest preferences, while frequent events such as views may include noise because guests might view listings without intending to book them. Thus, frequency-based selection can reduce cost while preserving the most valuable information.

\subsection{Feature Representation}
\label{sec:3:2}
We use a shared feature pool to organize all features, selecting relevant ones for each event, query, and listing to build feature representations. We have only used raw features without feature engineering to maintain the original semantics of the events. This feature pool encompasses three types of features:
\begin{itemize}[leftmargin=*]
    \item ID features - These include de-identified listing IDs, host IDs, and S2Cell IDs, which denote hierarchical geographical locations.
    \item Categorical features - These are discrete features with a limited and typically small set of possible values, such as country, listing room type, and the language of the Airbnb app.
    \item Continuous features - Continuous values like listing price.
\end{itemize}
A shared feature pool allows the same feature to be used across multiple contexts. For instance, a listing ID can be used in both listing candidates and specific events, such as listing view events, which can facilitate the learning of ID embeddings. In total, we only used less than a hundred features in \method{}, while our previous non-sequence model uses several hundreds of features.

\textbf{Processing categorical \& continuous features.} Handling categorical and continuous features is straightforward. Specifically, we encode each categorical feature using an embedding table and normalize continuous features with a batch normalization layer (and log transformation if necessary).

\textbf{Processing ID features.} Processing ID features is challenging due to their high cardinality. To reduce the modeling cost, we adopt Unified Embedding~\cite{coleman2023unified}. Specifically, we use a shared embedding table, \texttt{UnifiedEmb}, for all ID features. For each feature $x_{\text{id}}$, we use $m$ static random seeds, $\{s_1, s_2, ..., s_m\}$, and obtain its embedding by:
\begin{equation}
    \mathbf{e}_{x_\text{id}} = \texttt{Concat}_{i=1}^{m}\{\texttt{UnifiedEmb}(\texttt{Hash}_{s_i}(x_\text{id}))\},
    \label{eq:1}
\end{equation}
where $\texttt{Hash}_{s}(\cdot)$ hashes the input to a random row of \texttt{UnifiedEmb} with seed $s$, $\texttt{UnifiedEmb}(\cdot)$ performs an embedding lookup using the hashed ID, and \texttt{Concat} concatenates the retrieved embeddings. Similar to the original paper~\cite{coleman2023unified}, we found that given the same parameter budget, the unified embedding table is more scalable and performant than having separate id embeddings.

\textbf{Hierarchical S2Cell IDs.} S2Cell\footnote{\url{http://s2geometry.io/devguide/s2cell_hierarchy.html}} offers a hierarchical geographical indexing system. We employ S2Cell IDs from level 0 to level 13 (covering areas up to 1.27 $km^2$) to allow our model to understand location specifics at various scales. For instance, a listing in San Francisco is represented by 14 S2Cell IDs, capturing details from narrow streets to broader areas like the city or state. We find that this multi-level S2Cell ID approach enhances model performance compared to using a single S2Cell ID (see Section~\ref{subsection:alternative-q2}).

\subsection{Sequence Encoding}
\label{sec:3:3}

Given a guest sequence $\mathcal{S}_g = \{e_1, e_2, ..., e_t\}$, our goal is to encode it into a sequence embedding $\mathbf{e}_g$ to represent the guest's journey. We start by mapping the event features onto an event embedding space, resulting in $\{\mathbf{e}_i\}_{i=1}^t = \{\mathbf{e}_1, \mathbf{e}_2, ..., \mathbf{e}_t\}$, through a projection layer. Any sequence model can serve as the backbone of sequence encoder. Below, we describe our instantiation at Airbnb with self-attention.

\textbf{Self-attention sequence encoder.} Following~\cite{vaswani2017attention}, we encode the sequence using multi-head self-attention layers with position encoding and residual connections. Additionally, we apply a causal mask to prevent future events from influencing the encoding:
\begin{equation}
    \{\mathbf{e}^{\text{Pos}}_i\}_{i=1}^t = \texttt{Pos}(\{\mathbf{e}_i\}_{i=1}^t)),
    \label{eq:2}
\end{equation}
\begin{equation}
    \{\hat{\mathbf{e}}^{\text{Pos}}_i\}_{i=1}^t = \texttt{LayerNorm}(\texttt{MaskedAtt}(\{\mathbf{e}^{\text{Pos}}_i\}_{i=1}^t) + \{\mathbf{e}^{\text{Pos}}_i\}_{i=1}^t),
    \label{eq:3}
\end{equation}
where $\texttt{Pos}(\cdot)$ adds position encoding to the embeddings, $\texttt{MaskedAtt}(\cdot)$ applies multi-head self-attention with causal mask, and $\texttt{LayerNorm}(\cdot)$ means a layer normalization layer. We stack Eq.~\ref{eq:3} multiple times to obtain the final embeddings, and the final embedding corresponding to the last event is used as the sequence embedding $\mathbf{e}_g$.

\textbf{Why causal mask?} In Section~\ref{sec:3:5}, we will explain how using embeddings from events other than the last one can accelerate training speed. The causal mask guarantees that future events do not influence the embedding, preventing information leakage.

\textbf{Why exclude the feed forward layer?} Despite common uses along with self-attention layers~\cite{vaswani2017attention}, we find that feed forward layers lead to worse performances and slower training (Section \ref{subsection:alternative-q2}).

\subsection{Listing Ranking}
\label{sec:3:4}

Given a sequence embedding $\mathbf{e}_g$, a query $q$, and the set of retrieved listings $\mathcal{L}_q = \{l_1, l_2, ..., l_n\}$, our goal is to train a ranking model that scores each listing within $\mathcal{L}_q$. Specifically, at Airbnb, our goal is to rank listings with a higher booking likelihood higher.

\textbf{Label attribution.} As shown in Figure~\ref{fig:event_distribution}, guest view listings many times, leading to far more viewing events than booking events. In practice, many guests search for days to browse through listings. To maximally leverage booking labels, we attribute each booking to searches conducted in the previous 7 days. Specifically, if a listing in $\mathcal{L}_q$ is booked within the next 7 days, it is marked as a positive listing (label 1); otherwise, we assign label 0. We only include searches with at least one booking and sample up to four searches per day for each guest sequence to reduce data size while preserving diversity.

\textbf{Ranking model.} We concatenate the embeddings of the sequence, query, and listing and use an MLP to produce the listing's score. The choice of MLP is due to its simplicity and effectiveness. We will investigate more sophisticated architectures in future work.

\textbf{Loss.} We use listwise loss~\cite{cao2007learning} to train the embedding tables, sequence encoder, and the ranking model in an end-to-end manner. For $n$ listings in a search, let $z_i$ be the produced listing score, and $y_i$ be the label, where $i \in \{1, 2, ..., n\}$. The loss is computed as:
\begin{equation}
   \texttt{RankingLoss} = -\sum_{i=1}^n \frac{y_i}{\sum_{j=1}^n y_j} \log (\texttt{Softmax}(z_i)).
    \label{eq:4}
\end{equation}
The $\texttt{Softmax}(\cdot)$ denotes the Softmax operation. We normalize $y_i$ due to the possibility of multiple booked listings in a search.

\begin{figure}[t]
  \centering
  \includegraphics[width=0.46\textwidth]{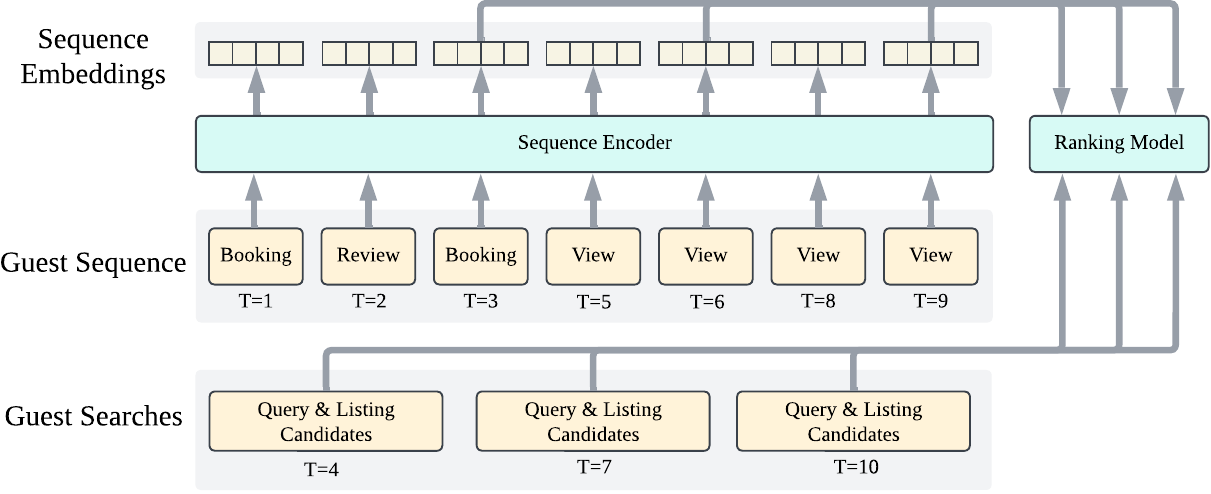}
  \caption{Batching of searches during training.}
  \vspace{-10pt}
  \label{fig:batch_search}
\end{figure}

\subsection{Efficient Training Strategies}
\label{sec:3:5}

To handle the large-scale production data, we have adopted several training strategies to improve training efficiency.

\textbf{Sequence bucketization.} Batching sequence data can easily cause data skewness issue because some guest journeys may have many events, whereas others have relatively few. To accommodate multiple journeys within a batch, we need to pad those with fewer events to match the journey with the highest number of events in the batch (280 for our model), which is inefficient because of the skewness. To address this issue, we bucketize sequences according to their lengths, using boundaries of 4, 16, 44, 88, 148, and 280. For example, a 15-event sequence is padded to meet the 16-event boundary. Then batches are drawn from these buckets, allowing for six possible batch sizes, improving the data handling efficiency. We implemented it with \texttt{bucket\_by\_sequence\_length} in Tensorflow.

\textbf{Batching of searches.} Recall that the encoder model provides not just the embedding of the last event but also the embeddings of all preceding events. In practice, these intermediate embeddings can serve as the sequence embeddings for searches immediately following them. As illustrated in Figure~\ref{fig:batch_search}, searches at a later time-stamp (T=4, T=7, and T=10) can utilize the embeddings from a previous time-stamp (T=3, T=6, and T=9). This allows us to process these three searches with just a single forward and backward pass of the encoder model, significantly reducing FLOPs since we often have a large sequence model and a relatively smaller ranking model.

\textbf{Sparse calculation of searches.} Similarly, batching searches requires padding those journeys with fewer searches and can lead to data skewness due to the significant variation in the number of searches per journey. Given that sequences are already bucketized by length, we can not use the same strategy for searches. To tackle this, we selectively input only the non-padded searches into the ranking model (e.g., using 
\texttt{tf.gather\_nd} in TensorFlow). For example, consider three journeys with 1, 2, and 3 searches, respectively. Instead of padding the first two journeys (resulting in 9 searches as input), we select only the 6 non-padded searches to pass into the ranking model. This trick can substantially improve training speed.

\begin{figure}[t]
  \centering
  \includegraphics[width=0.46\textwidth]{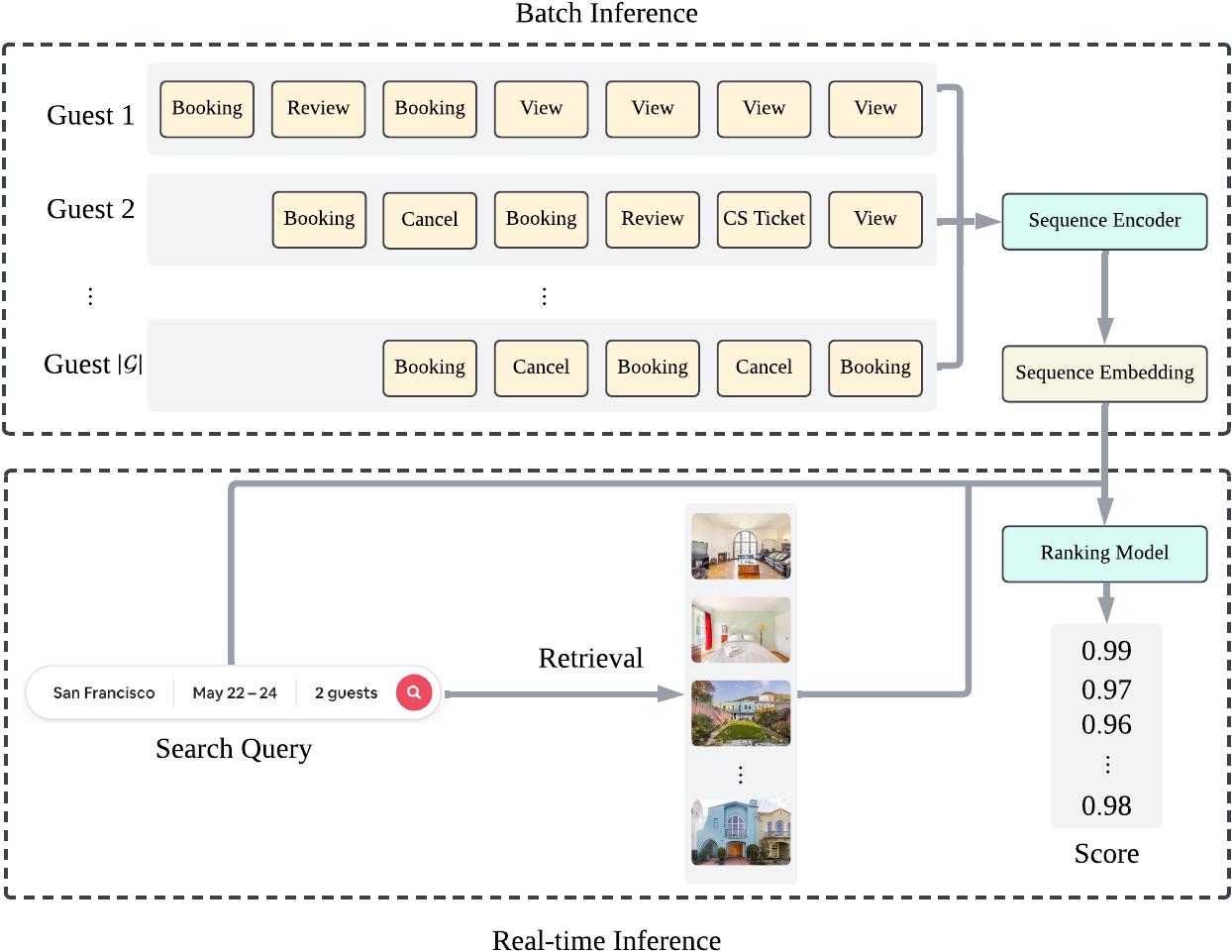}
  \caption{An illustration of model serving.}
  \vspace{-10pt}
  \label{fig:serving}
\end{figure}

\subsection{Model Serving}
\label{sec:3:6}

As illustrated in Figure~\ref{fig:serving}, our serving workflow comprises both batch and real-time inferences.

\textbf{Batch inference of sequence model.} Due to the quadratic increase in computational cost with the number of events, sequence models tend to be more expensive. Consequently, we employ a batch inference approach, periodically generating sequence embeddings for all guests and storing these embeddings in the database.

\textbf{Real-time inference of ranking model.} For the ranking model, we conduct real-time inference using the search query, retrieved listings, and the sequence embeddings.

\textbf{Discussion.} Using batch inference for the sequence model can also stabilize search results, potentially enhancing guest experience. This can mitigate some confusing scenarios where a listing visible in a previous search disappears shortly after in a subsequent search due to the too frequent updates of the sequence embedding.

\section{Experiments} \label{sec:4}

The experiments aim to answer the following questions: \textbf{Q1:} How does \method{} compare with the non-sequence-modeling baseline model? \textbf{Q2:} How will \method{} perform using alternative design choices?  \textbf{Q3:} Can the proposed training strategies improve efficiency? \textbf{Q4:} How does \method{} perform across different sequence lengths? \textbf{Q5:} Is \method{} sensitive to distribution shift across different days? \textbf{Q6:} Can the offline improvement of \method{} translate to final guest conversion online? 


\subsection{Experimental Setting}


\subsubsection{Production Baseline}

The baseline is Airbnb's previous production search ranking model, which relies on hand-crafted features rather than sequence modeling. At Airbnb, we adhered to a strict criterion for features that were launched into production, requiring each feature to demonstrate a gain in bookings or nights through online A/B testing. Over the past decades, we have incorporated several hundred features along with numerous model architecture improvements. As such, the baseline is very strong because it is the collective wisdom accumulated over many years of iterations.

\blue{The production baseline model consumes a rich set (around 900) of query, guest, host, listing, and price-related features that capture both short-term intent and long-term behavioral patterns. On the query side, we include basic request metadata (timing, device, locale, geography) as well as coarse signals of demand and search context. On the guest side, we incorporate aggregate statistics over past searches, views, and bookings to summarize engagement, and trip preferences, while avoiding any personally identifiable information. On the listing/host side, we use a combination of static attributes (capacity, room and property type, amenities, location, photos, and host profile characteristics) and dynamic marketplace signals (search impressions, page views, bookings, cancellations, review aggregates, and occupancy patterns), often aggregated at multiple spatial and temporal scales. Price features summarize both the current price breakdown and historical booking and exposure prices. Many of these signals are represented as low-dimensional embeddings (e.g., for categorical attributes and interaction buckets) or as compact aggregates (counts, rates, and moments), which allows the model to capture complex interactions between query, guest, host, and listing without exposing raw identifiers or fine-grained operational details.}

\blue{The production baseline uses a simple MLP architecture. In the past, we experimented with models that explicitly capture feature interactions, such as Deep \& Cross Networks (DCN)~\cite{wang2017deep}, but did not observe performance gains. A likely reason is that we have already invested heavily in feature engineering, so many interactions are encoded in the features themselves, leaving limited room for further improvement from more complex architectures.}

\blue{In contrast, \method{} uses fewer than 100 features in total, spanning guest, listing, host and journey signals. These are raw inputs; instead, all feature transformation and interaction modeling is delegated to the encoder.}

\subsubsection{Training Dataset}
\blue{The model is trained on booked guest sequences of 1 year. Each sequence includes non-frequent events from the past 7 years (long-term sequence), view events from the past 21 days (short-term sequence), and searches in the past 7 days (including the current date). We only use sequences that contain booking, which account for roughly 5\% of all the data. To control data scale and ensure quality, we apply the following filters, based on manual inspection and model training efficiency:

\begin{itemize}[leftmargin=*]
    \item Remove searches with fewer than 4 listing candidates.
    \item Keep at most 4 searches per day per guest.
    \item Keep only guest sequences with at least 1 search.
\end{itemize}

After filtering, the number of training sequences is in the magnitude of $10^8$. For each sequence, there are on average 6.9 searches. In terms of search–label pairs, the total count is around 900 millions.}


\subsubsection{Training Details}
Our model was trained with a training cluster. We used data from 9 days after the last date in training data for evaluation. For the sequence model, we used \blue{a standard 4-layer Transformer encoder with a model dimension of 128, positional encoding, and post-norm normalization}, with each input event being projected through an MLP of 1024-512-256-128; to reduce storage costs, the output embeddings were downscaled to a dimension of 32 via an MLP of 512-128-32. For the ranking model, we employed a light-weighted MLP of 1024-256-64-1 for efficient real-time inference. For model serving, our deployed models refresh sequence embeddings daily. There is potential to enhance online performance by updating these embeddings more frequently, such as every few hours. However, this needs additional infrastructure support in our side, and we  will investigate it in the future.

\subsubsection{Evaluation Protocol}

We evaluated our model using both the offline metric and online A/B testing. Offline, we measured performance with NDCG of booking labels. For each setting, we train three models and report their mean and standard deviations. For the online test, we focus on the following critical business metrics:

\begin{itemize}[leftmargin=*]
    \item Bookers - The number of guests who made at least one uncanceled booking. We used bookers rather than bookings to reduce the noise from some special guests, such as travel agents, who may book lots of listings in a short period of time.
    \item Booked Nights - The total number of uncanceled booked nights.
    \item Views - The total number of openings of a product listing page.
\end{itemize}

\blue{For more details, please refer to Appendix~\ref{appendix:A}.}

\begin{table}[t]
    \centering
    \small
    \caption{End-to-end offline NDCG difference in percent compared with the baseline model without sequence modeling. \method{}-Long, \method{}-Short and \method{}-Full correspond to using only long-term sequence, short-term sequence and both sequences respectively}
    \label{tab:offline}
    \begin{tabular}{c|c}
    \toprule
    \textbf{Model} & \textbf{Offline NDCG} \\
    \midrule
     Baseline (no sequence modeling) & $\;\;0.00\% \pm 0.01\%$ \\
     \method{}-Long & $+0.44\% \pm 0.01\%$ \\
     \method{}-Short & $+1.30\% \pm 0.03\%$ \\
     \method{}-Full & $+1.48\% \pm 0.03\%$ \\
    \bottomrule
    \end{tabular}
\end{table}

\subsection{Comparison with Baseline (Q1)}

Table~\ref{tab:offline} presents the NDCG improvement relative to the baseline for three variants of \method{}: using long-term, short-term, and both sequences, respectively. Our observations are as follows: (i) All three models significantly outperform the baseline, demonstrating the effectiveness of sequence modeling. It is noteworthy that at Airbnb, a 0.3\% improvement in NDCG is considered significant given the highly optimized nature of our ranking model. (ii) The short-term sequence contributes more to the offline NDCG improvement. A possible reason is that the final booking is likely to be from listings viewed recently, making the short-term sequence with certain view events a strong indicator of booking intent. 

\begin{figure}[t]
  \centering
  \includegraphics[width=0.46\textwidth]{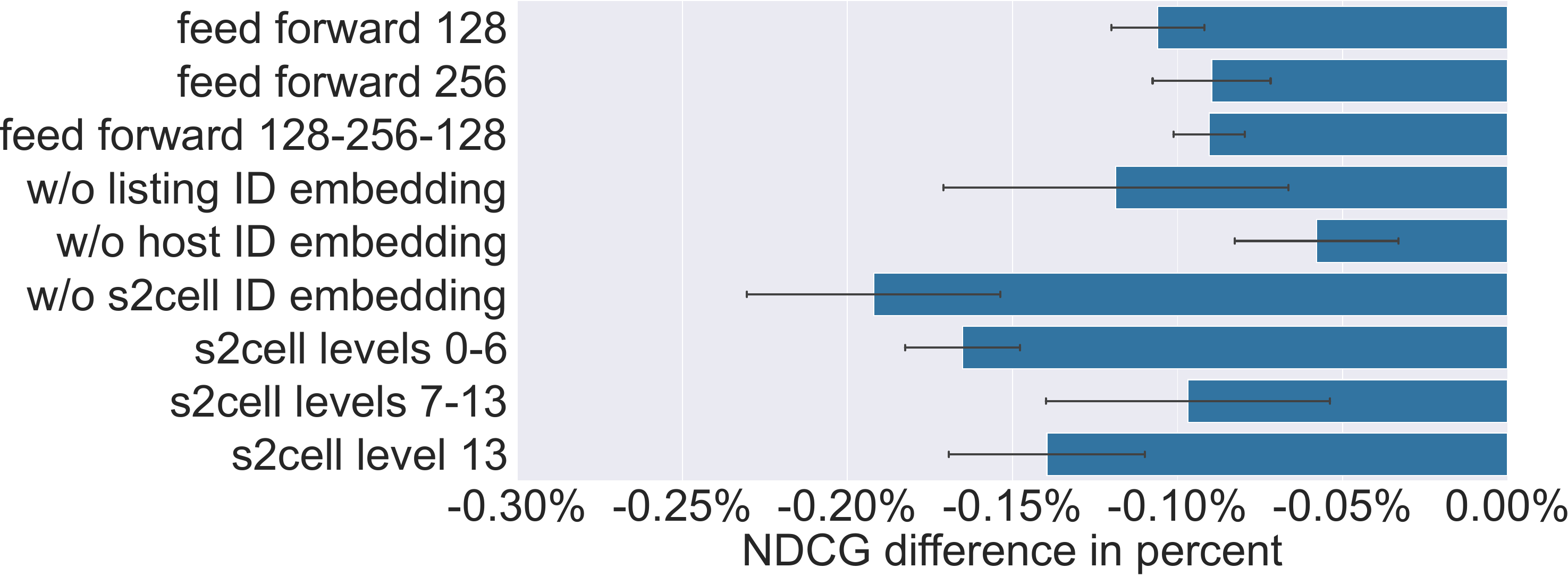}
  \vspace{-10pt}
  \caption{NDCG of alternative design choices. We report the percentage difference compared to \method{}-Full.}
  \label{fig:alternative}
  \vspace{-10pt}
\end{figure}

\subsection{Analysis of Alternative Design Choices (Q2)}
\label{subsection:alternative-q2}

In Figure~\ref{fig:alternative}, we test several alternative design choices.

\textbf{Is a feed forward layer helpful (the bar at 1-3)?} No. Recall that we have excluded the feed forward layer in the sequence encoder. Here, we test three different configurations of feed forward layers with sizes of 128, 256, and 128-256-128, respectively. We observe a drop in NDCG across all configurations. Thus, we excluded feed forward layer to reduce model parameters. We suspect this is because our MLP already maps event features to embeddings, making the feed-forward layer in the sequence encoder redundant. The underlying causes will be studied in our future work.

\textbf{Are ID embeddings helpful (the bar at 4-6)?} Yes. We individually remove listing ID, host ID, and S2Cell ID to study their impact and observe a decrease in NDCG (from 0.05\% to 0.2\%) for each case. Notably, removing S2Cell ID results in the largest NDCG drop. This suggests that the rich location information from S2Cell ID provided by S2Cell ID plays a crucial role in enhancing booking predictions. This is expected and unique to the Airbnb platform, as guests are often sensitive to location specifics of listings.


\textbf{Are hierarchical S2Cell IDs helpful (the bar at 7-9)?} Yes. Recall that we used 14 levels of S2Cell IDs, ranging from level 0 (broad areas like continents) to level 13 (narrow areas like street segments). To understand whether all 14 levels are necessary, we test three ablations: levels 0-6 (broader areas), levels 7-13 (narrower areas), and level 13 alone. (i) NDCG decreases across all ablations. (ii) Although level 13 theoretically contains all location information from levels 0-13 (e.g., an S2Cell ID for a specific street area in San Francisco should also include city and state details), using it alone underperforms compared to using multiple S2Cell levels. One possible reason is that the model may better understand locations using S2Cell inputs of varying granularities. These observations suggest that utilizing all 14 hierarchical S2Cell IDs can enhance the model's ability to learn location representations.


\subsection{Analysis of Acceleration Strategies (Q3)}

\begin{table}[t]
    \centering
    \small
    \caption{Training throughput when removing each training acceleration strategy compared to \method{}-Full.}
    \label{tab:accelerating}
    \begin{tabular}{c|c}
    \toprule
    \textbf{Ablation} & \textbf{Training throughput difference} \\
    \midrule
     w/o batching of searches & $-75.41\% \pm 0.53\%$ \\
     w/o sequence bucketization & $-29.35\% \pm 1.42\%$ \\
     w/o sparse calculation & $-20.54\% \pm 0.63\%$ \\
    \bottomrule
    \end{tabular}
\end{table}

Table~\ref{tab:accelerating} shows the impact of removing each of the training acceleration strategies discussed in Section~\ref{sec:3:5}. We find that each strategy significantly increases training throughput. Notably, removing the batching of searches reduces throughput to approximately 1/4 of its original rate. A small trick for sparse calculation also enhances training efficiency notably. These acceleration strategies are crucial for efficiently training our model on large-scale production data.

\subsection{Analysis of Sequence Lengths (Q4)}

\begin{figure}[t]
  \centering

  \begin{subfigure}[b]{0.232\textwidth}
    \centering
    \includegraphics[width=0.99\textwidth]{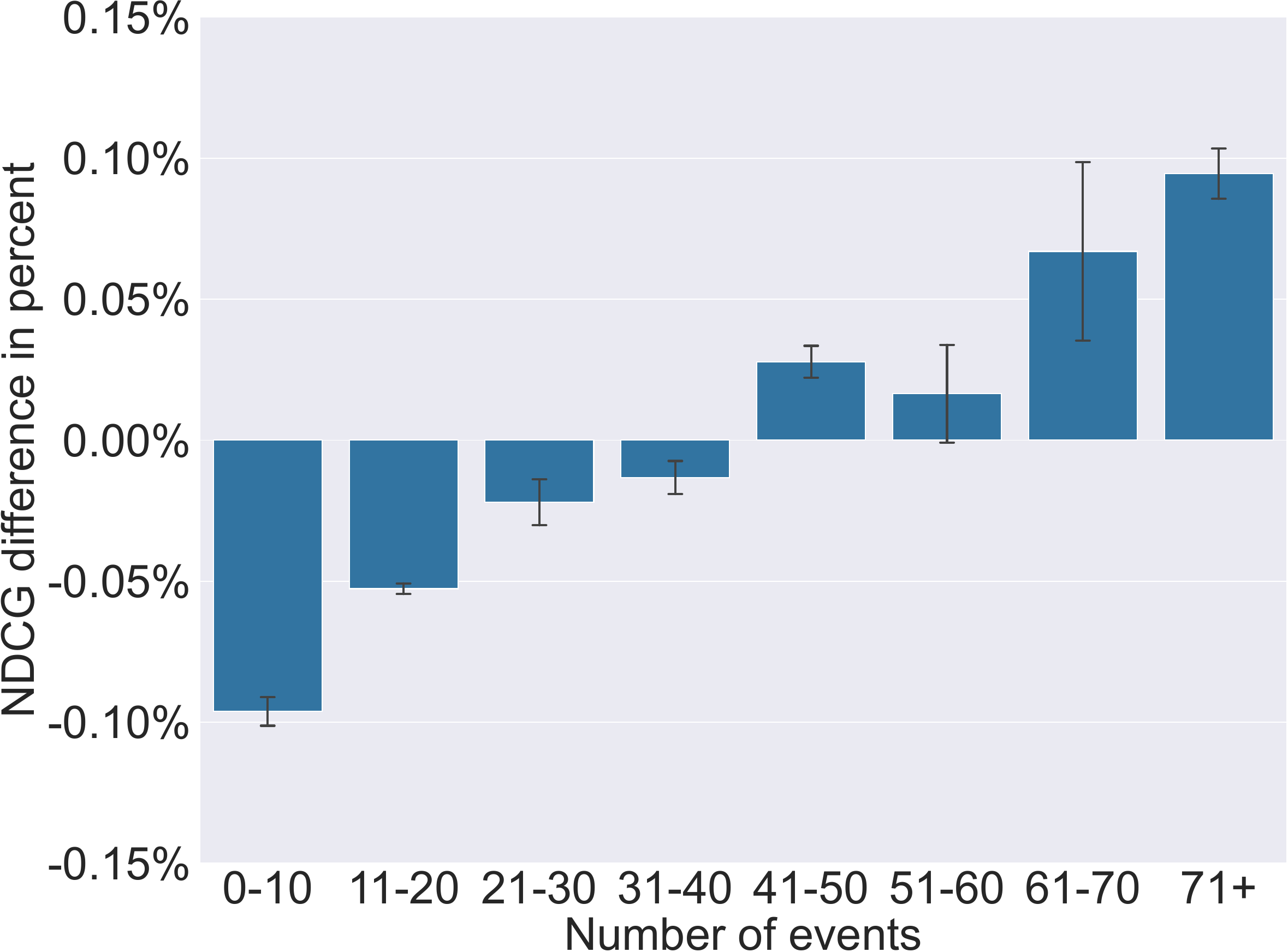}
  \end{subfigure}%
  \begin{subfigure}[b]{0.228\textwidth}
    \centering
    \includegraphics[width=0.99\textwidth]{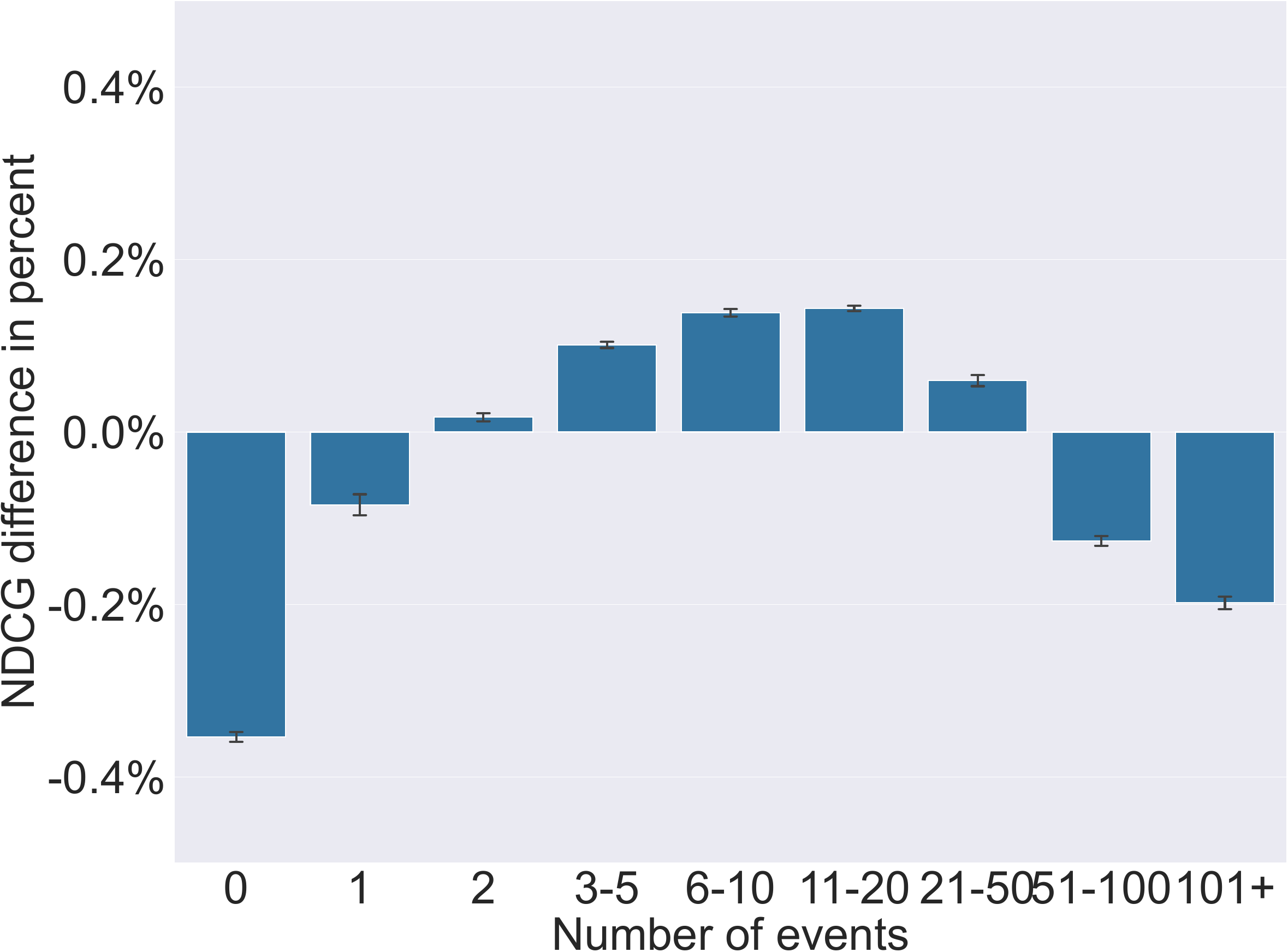}
  \end{subfigure}%
  \caption{NDCG w.r.t. the number of long-term (left) and short-term (right) events. We report the percentage difference compared to the overall NDCG.}
  \label{fig:seq_length}
\end{figure}

To  glean some insights of model's behavior, we segment the evaluation dataset based on long-term and short-term sequence lengths and report the NDCG in Figure~\ref{fig:seq_length}. We observe distinct patterns between long-term and short-term sequences. Specifically, for long-term sequences, there is a clear trend: the NDCG increases with sequence length, aligning with the expectation that more long-term events (e.g., bookings and reviews) enhances guest intention prediction. Interestingly, short-term sequences exhibit peak NDCG with 11-20 events, after which the NDCG notably decline with more events. A possible explanation is that predicting the intentions of guests who view many listings becomes challenging, as these guests may browse extensively without any concrete intent to book, aligning with our hypothesis that view events are noisy. This finding also validates the efficacy of our frequency-based selection strategy.

\begin{figure}[t]
  \centering
  \includegraphics[width=0.46\textwidth]{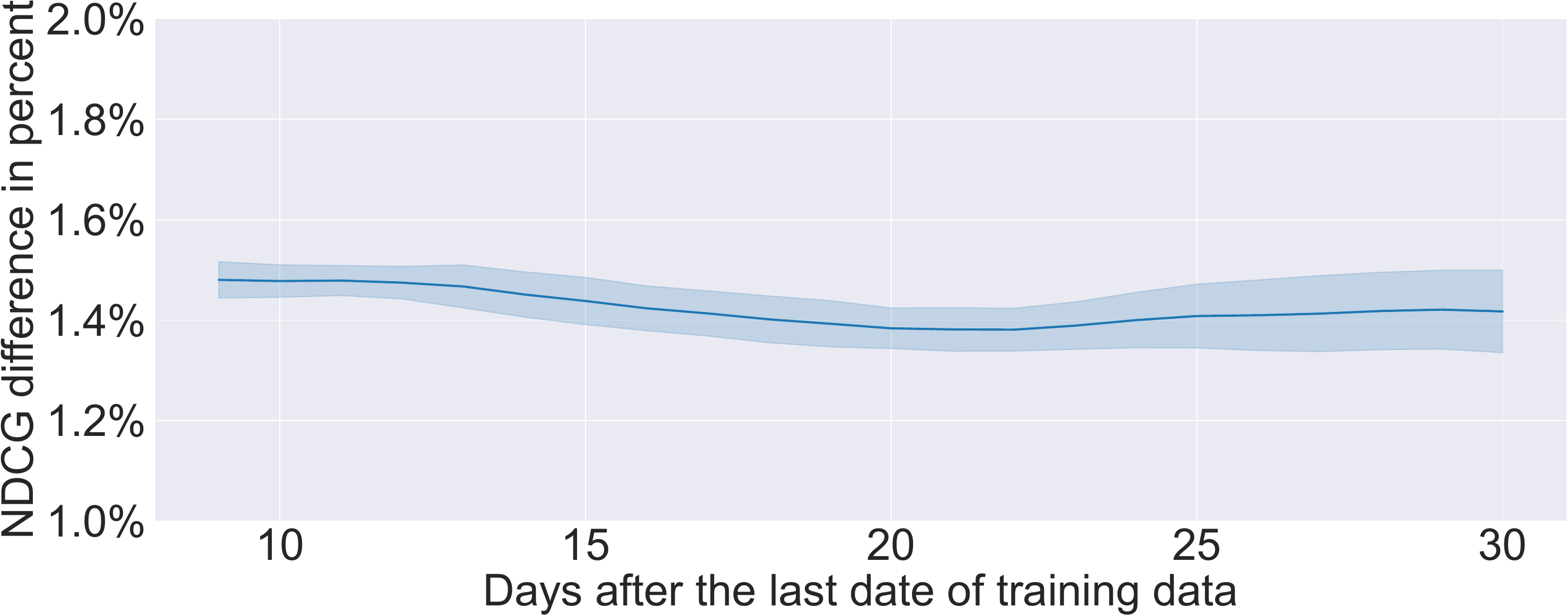}
  \caption{NDCG evaluated up to 1 month after the last date of the training data. We report the the percentage difference compared to the baseline model without sequence modeling. The shaded area represents mean $\pm$ standard deviation.}
  \label{fig:distribution}
\end{figure}

\subsection{Sensitivity to Distribution Shift (Q5)}

We investigate the sensitivity of our model to shifts in data distribution. In Figure~\ref{fig:distribution}, we report the relative NDCG improvement of \method{} compared to the baseline on evaluation data extending up to one month beyond the training data's last date. We observe that \method{} consistently outperforms the baseline by a large margin (>1.4\% for majority of the dates) with only a minor decrease (<0.1\%) in NDCG over this period. This indicates that frequent retraining of the model is unnecessary, which is a desirable property for deployment in production environments.

\subsection{Online Results (Q6)}
\label{sec:4:7}

Offline NDCG tends to be biased towards bookers since our evaluation dataset only includes guests with booking labels, despite they representing only a small fraction of overall traffic. However, our primary goal is to convert non-bookers into bookers, which is hard to assess directly with offline data. 

To test whether the improvement of \method{} can generalize to non-bookers, we conducted online A/B experiments over a three-week period to compare \method{}-Long and \method{}-Full with the baseline, reported in Table~\ref{tab:online}. The results show significant increases for all the metrics, highlighting the benefits of incorporating both long-term and short-term sequences. Interestingly, short-term sequences bring more gains in booked nights and views than bookers. A possible explanation is that guests planning to book more nights tend to be more cautious and hesitant, leading them to explore a wider range of listings before finalizing their decision. By learning from the recent viewing activity, our model could help these hesitant guests in identifying their desired listings, ultimately leading to more booked nights.

To further test whether \method{} can truly identify guest needs, we applied \method{}-Full to recommend relevant listings to guests in promotional emails. Similar to search ranking, email listing ranking has been iterated for many years with strict launching criteria—a model can only be launched if it demonstrates significant gains in online A/B testing. As such, the baseline is very strong. Table~\ref{tab:online2} shows the results. \method{}-Full significantly increases the likelihood of users clicking into the promotional email by more than 5\%, leading to more bookers and booked nights.

\section{Discussion} \label{sec:discuss}
We discuss several promising directions we explored but ultimately did not pursue, in addition to those covered in Section~\ref{subsection:alternative-q2}. The rationale for not adopting these approaches is to avoid added complexity or due to the gap between online and offline metrics. \blue{We provide additional discussion and results in Appendix~\ref{appendix:B}.}

\begin{table}[t]
    \centering
    \small
    \caption{Online A/B testing results compared with the previous production model, conducted over a three-week period. * indicates p < 0.05, while ** means p < 0.01. Note that the production model in Airbnb search ranking has been already iterated for many years. A 0.3\% improvement is considered very significant (see our previous work~\cite{tan2023optimizing,tang2024multi,haldar2019applying}).  }
    \label{tab:online}
    \begin{tabular}{c|c|c|c}
    \toprule
    \textbf{Model} & \textbf{Bookers} & \textbf{Booked Nights} & \textbf{Views} \\
    \midrule
     \method{}-Long & $+0.31\%$*$\,\:$ & $+0.28\%$$\,\:\,\:$ & $+0.38\%$** \\
     \method{}-Full & $+0.55\%$** & $+0.82\%$** & $+0.90\%$** \\
    \bottomrule
    \end{tabular}
\end{table}

\begin{table}[t]
    \centering
    \small
    \caption{Online A/B testing results compared with the previous production model in email listing ranking. Views suggest the email is clicked by the guest. * indicates p < 0.05, while ** means p < 0.01. }
    \label{tab:online2}
    \begin{tabular}{c|c|c|c}
    \toprule
    \textbf{Model} & \textbf{Bookers} & \textbf{Booked Nights} & \textbf{Views} \\
    \midrule
     \method{}-Full & $+0.16\%$*$\,\:$ & $+0.23\%$*$\,\:$ & $+5.04\%$** \\
    \bottomrule
    \end{tabular}
    \vspace{-10pt}
\end{table}

\begin{itemize}[leftmargin=*]
\item \textbf{Larger or more sophisticated models.} We have tried increasing the number of parameters by expanding embedding table size, adding more layers, or increasing the embedding dimension. We have also explored various model architecture and tuning tweaks, such as different approaches to modeling the interactions among query, listing, and sequence embeddings, as well as different learning rate schedulers. While we did observe some offline improvements (which may also be difficult to translate to online metric gains), the improvements were marginal compared to the costs they introduced. They also increased code complexity, accumulating technical debt. In the end, we kept only the components that worked and maintained a simple model architecture.
\item \textbf{Real-time inference of sequence model.} Ignoring serving costs, this approach initially seemed very promising as it yields significant gains in offline ranking metrics compared to daily inference (more than 5\% improvement). However, we found that this offline gain does not translate to online gain due to retargeting, where the ranker repeatedly shows listings that users viewed previously. This typically does not help guests make booking decisions, since they are already aware of the retargeted listings—despite the fact that it can significantly improve ranking metrics. This phenomenon differs from social networks, where repeatedly showing content is likely to enhance the likelihood of user engagement. Booking is often a decision involving hundreds of dollars; guests will not book a listing simply because it is shown many times. Rather, our ranking system should help guests identify their needs instead of retargeting.
\item \textbf{Adding search events into the sequence.} Another very frequent event in search ranking is search itself, which could reveal guest intent. However, in our initial trial, adding search events was not helpful, possibly due to the noisiness of search data.

\item \textbf{Multi-task learning.} In addition to booking labels, other guest actions can naturally be used as labels, such as views. This naturally leads to multi-task learning, where we predict both booking and view labels. We observed some decent gains in offline metrics with this approach. However, we found that these gains do not translate to online improvements. A possible reason is that view events are quite noisy—guests may view a listing simply for comparison purposes. Using views as labels does not help improve bookings. While this was our initial finding, we believe there is still potential and will continue exploring this direction in the future. From a data perspective, this approach is promising since we can leverage significantly more training data from sessions where guests only viewed but did not book. 

\end{itemize}

\begin{table}[t]
    \centering
    \small
    \caption{Offline NDCG of DIN-style interaction over \method{}.}
    \vspace{-10pt}
    \label{tab:din}
    \begin{tabular}{c|c}
    \toprule
    \textbf{Model} & \textbf{Offline NDCG} \\
    \midrule
     DIN-style \method{} & $+0.20\% \pm 0.01\%$ \\
    \bottomrule
    \end{tabular}
    \vspace{-10pt}
\end{table}

\section{Related Work} \label{sec:5}

\textbf{Sequence modeling for recommendation \& ranking.} Modeling user behaviors with sequence modeling has gained increasing popularity in recommendation and ranking algorithms. Initial efforts in capturing user intentions involve the use of Markov chains~\cite{shani2005mdp,salakhutdinov2007restricted,he2017translation,he2016fusing}. With the emergence of deep learning, Recurrent neural networks like LSTM~\cite{graves2012long} and GRU~\cite{chung2014empirical} have been extensively studied for recommendation~\cite{li2017neural,quadrana2017personalizing,hidasi2015session,yu2016dynamic,donkers2017sequential}. Concurrently, other neural architectures, such as convolutional networks~\cite{tang2018personalized} and memory networks~\cite{chen2018sequential}, have also been used for sequence modeling. More recently, inspired by the remarkable success of self-attention in NLP~\cite{vaswani2017attention,radford2019language,brown2020language,devlin2019bert,touvron2023llama}, Transformer-based sequence models have demonstrated impressive performance in recommendation~\cite{sun2019bert4rec,kang2018self,wu2020sse,li2020time,chen2022intent,fan2022sequential,du2023frequency,zhang2023adaptive,zhou2023attention}. The existing research has mainly focused on recommendation without search queries, where user intent is primarily inferred from the user sequence. However, in search ranking scenarios, user intent is mainly conveyed through the queries they submit. Our work demonstrates that in search ranking platforms like Airbnb, sequence modeling can still play a crucial role as a complementary signal alongside queries. Furthermore, beyond the algorithmic and neural architecture enhancements explored in prior research, we have emphasized and implemented practical design considerations for production, successfully deploying and testing our model with real guest traffic.

\blue{\textbf{Relationship to other sequence models.} \method{} is closely related to other sequential recommendation models~\cite{kang2018self,sun2019bert4rec,zhou2018deep} and here, we discuss its relationship to several popular variants.

SASRec (Self-Attentive Sequential Recommendation)~\cite{kang2018self} uses a self-attention network to model user action sequences and predict the next action. \method{} adopts a similar approach to encode guest actions. The key differences are that \method{} uses a sequence consisting of heterogeneous event types and ranks listings within a candidate set instead of predicting the next item. We also introduce training strategies tailored to search ranking. In this sense, \method{} can be viewed as an adaptation of SASRec to the search ranking setting.

BERT4Rec~\cite{sun2019bert4rec} is similar to SASRec but uses a bidirectional encoder. While bidirectional encoding can theoretically better capture user sequences by allowing tokens to attend in both directions, we found it to be neutral for our particular application empirically. Furthermore, in industrial settings, modest architectural gains are often outweighed by training speed (e.g. batched-search acceleration in \method{}), especially at our data scale. For this reason, we did not use a bidirectional encoder in \method{}.

DIN (Deep Interest Network)~\cite{zhou2018deep} is another sequential recommendation model, originally proposed for click-through rate prediction. The main difference is that DIN is target-aware: it explicitly models interactions between the target item and each event in the sequence. In contrast, \method{} only models interactions between target items and the user embedding output by the encoder. DIN-style interaction is an “early-fusion” approach and can be more expressive, whereas \method{} is closer to “late-fusion.” However, early-fusion is much more expensive at serving time because we cannot precompute encoder outputs offline; \method{} can precompute user embeddings offline (where the encoder is expensive) and use a lightweight ranking head online.

Despite these serving challenges, we are interested in whether DIN-style interactions can improve offline performance. To test this idea, we project query and candidate features into the same representation space as other events and insert these “search events” into the sequence, allowing them to interact via self-attention. This does not exactly replicate DIN but approximates early fusion with minimal changes to data and model, and still allows us to reuse acceleration strategies such as batched searches. We call this variant DIN-style \method{}. Trained on the same number of sequences, it takes about 15\% more time due to the longer sequences. As shown in Table~\ref{tab:din}, we observe a decent offline NDCG gain of $+0.20\%$, even without hyperparameter tuning. We plan to further tune hyperparameters and investigate ways to reduce serving latency for this variant in future work.}

\textbf{Search ranking in industry.} Search ranking is a key function in various search platforms including Google~\cite{ziakis2019important}, Amazon~\cite{buyl2023rankformer,missault2021addressing,luo2022query}, eBay~\cite{trotman2017architecture}, Walmart~\cite{magnani2022semantic}, Pinterest~\cite{pancha2022pinnerformer,Xia2023TransActTR} and Airbnb~\cite{tan2023optimizing,haldar2023learning,tang2024multi,haldar2019applying,abdool2020managing}. Most of these solutions rely on hand-crafted features to represent user sequences. The Generative Recommender from Meta~\cite{Zhai2024ActionsSL} introduces an industrial-scale sequence model, which is designed for recommendation rather than search ranking. Pinterest also utilizes Transformer to learn user embeddings~\cite{pancha2022pinnerformer,Xia2023TransActTR}. The embeddings are then served as features in ranking models. Our work differs significantly from theirs, as our model focuses on booking conversion rather than engagement. As a result, engagement signals—such as listing views—often reflect exploratory behavior and can be quite noisy, making it challenging to infer true booking intent. Additionally, the complexity of guest events differ greatly from those on social media platforms.

\section{Conclusions and Future Work} \label{sec:6}

In this work, we introduce \method{}, a sequence modeling solution for encoding Airbnb guest journeys. Deploying sequence models at Airbnb presents unique challenges, and we discuss how we address these production hurdles, including data selection, label attribution, and training and model serving considerations. Through extensive offline and online A/B testing, we demonstrate that \method{} significantly boosts key business metrics, helping guests find what they want and enabling hosts to more efficiently match with guests. In the future, in addition to the directions in Section~\ref{sec:discuss}, we plan to enhance our serving infrastructure to allow sequence embeddings to be updated every few hours, and incorporate additional guest events into the sequence, such as wishlist creations, payment page visits, and map movements. Finally, we plan to explore reinforcement learning with sequence model.

\bibliographystyle{ACM-Reference-Format}
\bibliography{ref}

@inproceedings{tan2023optimizing,
  title={Optimizing airbnb search journey with multi-task learning},
  author={Tan, Chun How and Chan, Austin and Haldar, Malay and Tang, Jie and Liu, Xin and Abdool, Mustafa and Gao, Huiji and He, Liwei and Katariya, Sanjeev},
  booktitle={Proceedings of the 29th ACM SIGKDD Conference on Knowledge Discovery and Data Mining},
  pages={4872--4881},
  year={2023}
}

@inproceedings{haldar2023learning,
  title={Learning To Rank Diversely At Airbnb},
  author={Haldar, Malay and Abdool, Mustafa and He, Liwei and Davis, Dillon and Gao, Huiji and Katariya, Sanjeev},
  booktitle={Proceedings of the 32nd ACM International Conference on Information and Knowledge Management},
  pages={4609--4615},
  year={2023}
}

@inproceedings{tang2024multi,
  title={Multi-objective Learning to Rank by Model Distillation},
  author={Tang, Jie and Gao, Huiji and He, Liwei and Katariya, Sanjeev},
  booktitle={Proceedings of the 30th ACM SIGKDD Conference on Knowledge Discovery and Data Mining},
  pages={5783--5792},
  year={2024}
}

@inproceedings{haldar2019applying,
  title={Applying deep learning to airbnb search},
  author={Haldar, Malay and Abdool, Mustafa and Ramanathan, Prashant and Xu, Tao and Yang, Shulin and Duan, Huizhong and Zhang, Qing and Barrow-Williams, Nick and Turnbull, Bradley C and Collins, Brendan M and others},
  booktitle={proceedings of the 25th ACM SIGKDD international conference on knowledge discovery \& Data Mining},
  pages={1927--1935},
  year={2019}
}

@inproceedings{abdool2020managing,
  title={Managing diversity in airbnb search},
  author={Abdool, Mustafa and Haldar, Malay and Ramanathan, Prashant and Sax, Tyler and Zhang, Lanbo and Manaswala, Aamir and Yang, Lynn and Turnbull, Bradley and Zhang, Qing and Legrand, Thomas},
  booktitle={Proceedings of the 26th ACM SIGKDD International Conference on Knowledge Discovery \& Data Mining},
  pages={2952--2960},
  year={2020}
}

@article{coleman2023unified,
  title={Unified Embedding: Battle-tested feature representations for web-scale ML systems},
  author={Coleman, Benjamin and Kang, Wang-Cheng and Fahrbach, Matthew and Wang, Ruoxi and Hong, Lichan and Chi, Ed and Cheng, Derek},
  journal={Advances in Neural Information Processing Systems},
  volume={36},
  pages={56234--56255},
  year={2023}
}

@article{vaswani2017attention,
  title={Attention is all you need},
  author={Vaswani, Ashish and Shazeer, Noam and Parmar, Niki and Uszkoreit, Jakob and Jones, Llion and Gomez, Aidan N and Kaiser, {\L}ukasz and Polosukhin, Illia},
  journal={Advances in neural information processing systems},
  volume={30},
  year={2017}
}

@inproceedings{cao2007learning,
  title={Learning to rank: from pairwise approach to listwise approach},
  author={Cao, Zhe and Qin, Tao and Liu, Tie-Yan and Tsai, Ming-Feng and Li, Hang},
  booktitle={Proceedings of the 24th international conference on Machine learning},
  pages={129--136},
  year={2007}
}

@article{radford2019language,
  title={Language models are unsupervised multitask learners},
  author={Radford, Alec and Wu, Jeffrey and Child, Rewon and Luan, David and Amodei, Dario and Sutskever, Ilya and others},
  journal={OpenAI blog},
  volume={1},
  number={8},
  pages={9},
  year={2019}
}

@article{brown2020language,
  title={Language models are few-shot learners},
  author={Brown, Tom and Mann, Benjamin and Ryder, Nick and Subbiah, Melanie and Kaplan, Jared D and Dhariwal, Prafulla and Neelakantan, Arvind and Shyam, Pranav and Sastry, Girish and Askell, Amanda and others},
  journal={Advances in neural information processing systems},
  volume={33},
  pages={1877--1901},
  year={2020}
}

@inproceedings{devlin2019bert,
  title={Bert: Pre-training of deep bidirectional transformers for language understanding},
  author={Devlin, Jacob and Chang, Ming-Wei and Lee, Kenton and Toutanova, Kristina},
  booktitle={Proceedings of the 2019 conference of the North American chapter of the association for computational linguistics: human language technologies, volume 1 (long and short papers)},
  pages={4171--4186},
  year={2019}
}

@article{touvron2023llama,
  title={Llama: Open and efficient foundation language models},
  author={Touvron, Hugo and Lavril, Thibaut and Izacard, Gautier and Martinet, Xavier and Lachaux, Marie-Anne and Lacroix, Timoth{\'e}e and Rozi{\`e}re, Baptiste and Goyal, Naman and Hambro, Eric and Azhar, Faisal and others},
  journal={arXiv preprint arXiv:2302.13971},
  year={2023}
}

@article{shani2005mdp,
  title={An MDP-based recommender system},
  author={Shani, Guy and Heckerman, David and Brafman, Ronen I},
  journal={Journal of machine Learning research},
  volume={6},
  number={Sep},
  pages={1265--1295},
  year={2005}
}

@inproceedings{salakhutdinov2007restricted,
  title={Restricted Boltzmann machines for collaborative filtering},
  author={Salakhutdinov, Ruslan and Mnih, Andriy and Hinton, Geoffrey},
  booktitle={Proceedings of the 24th international conference on Machine learning},
  pages={791--798},
  year={2007}
}

@inproceedings{he2017translation,
  title={Translation-based recommendation},
  author={He, Ruining and Kang, Wang-Cheng and McAuley, Julian},
  booktitle={Proceedings of the eleventh ACM conference on recommender systems},
  pages={161--169},
  year={2017}
}

@inproceedings{he2016fusing,
  title={Fusing similarity models with markov chains for sparse sequential recommendation},
  author={He, Ruining and McAuley, Julian},
  booktitle={2016 IEEE 16th international conference on data mining (ICDM)},
  pages={191--200},
  year={2016},
  organization={IEEE}
}

@inproceedings{donkers2017sequential,
  title={Sequential user-based recurrent neural network recommendations},
  author={Donkers, Tim and Loepp, Benedikt and Ziegler, J{\"u}rgen},
  booktitle={Proceedings of the eleventh ACM conference on recommender systems},
  pages={152--160},
  year={2017}
}

@inproceedings{li2017neural,
  title={Neural attentive session-based recommendation},
  author={Li, Jing and Ren, Pengjie and Chen, Zhumin and Ren, Zhaochun and Lian, Tao and Ma, Jun},
  booktitle={Proceedings of the 2017 ACM on Conference on Information and Knowledge Management},
  pages={1419--1428},
  year={2017}
}

@inproceedings{quadrana2017personalizing,
  title={Personalizing session-based recommendations with hierarchical recurrent neural networks},
  author={Quadrana, Massimo and Karatzoglou, Alexandros and Hidasi, Bal{\'a}zs and Cremonesi, Paolo},
  booktitle={proceedings of the Eleventh ACM Conference on Recommender Systems},
  pages={130--137},
  year={2017}
}

@article{hidasi2015session,
  title={Session-based recommendations with recurrent neural networks},
  author={Hidasi, Bal{\'a}zs and Karatzoglou, Alexandros and Baltrunas, Linas and Tikk, Domonkos},
  journal={arXiv preprint arXiv:1511.06939},
  year={2015}
}

@inproceedings{yu2016dynamic,
  title={A dynamic recurrent model for next basket recommendation},
  author={Yu, Feng and Liu, Qiang and Wu, Shu and Wang, Liang and Tan, Tieniu},
  booktitle={Proceedings of the 39th International ACM SIGIR conference on Research and Development in Information Retrieval},
  pages={729--732},
  year={2016}
}

@article{graves2012long,
  title={Long short-term memory},
  author={Graves, Alex and Graves, Alex},
  journal={Supervised sequence labelling with recurrent neural networks},
  pages={37--45},
  year={2012},
  publisher={Springer}
}

@article{chung2014empirical,
  title={Empirical evaluation of gated recurrent neural networks on sequence modeling},
  author={Chung, Junyoung and Gulcehre, Caglar and Cho, KyungHyun and Bengio, Yoshua},
  journal={arXiv preprint arXiv:1412.3555},
  year={2014}
}

@inproceedings{tang2018personalized,
  title={Personalized top-n sequential recommendation via convolutional sequence embedding},
  author={Tang, Jiaxi and Wang, Ke},
  booktitle={Proceedings of the eleventh ACM international conference on web search and data mining},
  pages={565--573},
  year={2018}
}

@inproceedings{chen2018sequential,
  title={Sequential recommendation with user memory networks},
  author={Chen, Xu and Xu, Hongteng and Zhang, Yongfeng and Tang, Jiaxi and Cao, Yixin and Qin, Zheng and Zha, Hongyuan},
  booktitle={Proceedings of the eleventh ACM international conference on web search and data mining},
  pages={108--116},
  year={2018}
}

@inproceedings{sun2019bert4rec,
  title={BERT4Rec: Sequential recommendation with bidirectional encoder representations from transformer},
  author={Sun, Fei and Liu, Jun and Wu, Jian and Pei, Changhua and Lin, Xiao and Ou, Wenwu and Jiang, Peng},
  booktitle={Proceedings of the 28th ACM international conference on information and knowledge management},
  pages={1441--1450},
  year={2019}
}

@inproceedings{kang2018self,
  title={Self-attentive sequential recommendation},
  author={Kang, Wang-Cheng and McAuley, Julian},
  booktitle={2018 IEEE international conference on data mining (ICDM)},
  pages={197--206},
  year={2018},
  organization={IEEE}
}

@inproceedings{wu2020sse,
  title={SSE-PT: Sequential recommendation via personalized transformer},
  author={Wu, Liwei and Li, Shuqing and Hsieh, Cho-Jui and Sharpnack, James},
  booktitle={Proceedings of the 14th ACM conference on recommender systems},
  pages={328--337},
  year={2020}
}

@inproceedings{li2020time,
  title={Time interval aware self-attention for sequential recommendation},
  author={Li, Jiacheng and Wang, Yujie and McAuley, Julian},
  booktitle={Proceedings of the 13th international conference on web search and data mining},
  pages={322--330},
  year={2020}
}

@inproceedings{chen2022intent,
  title={Intent contrastive learning for sequential recommendation},
  author={Chen, Yongjun and Liu, Zhiwei and Li, Jia and McAuley, Julian and Xiong, Caiming},
  booktitle={Proceedings of the ACM web conference 2022},
  pages={2172--2182},
  year={2022}
}

@inproceedings{fan2022sequential,
  title={Sequential recommendation via stochastic self-attention},
  author={Fan, Ziwei and Liu, Zhiwei and Wang, Yu and Wang, Alice and Nazari, Zahra and Zheng, Lei and Peng, Hao and Yu, Philip S},
  booktitle={Proceedings of the ACM web conference 2022},
  pages={2036--2047},
  year={2022}
}

@inproceedings{du2023frequency,
  title={Frequency enhanced hybrid attention network for sequential recommendation},
  author={Du, Xinyu and Yuan, Huanhuan and Zhao, Pengpeng and Qu, Jianfeng and Zhuang, Fuzhen and Liu, Guanfeng and Liu, Yanchi and Sheng, Victor S},
  booktitle={Proceedings of the 46th International ACM SIGIR conference on research and development in information retrieval},
  pages={78--88},
  year={2023}
}

@inproceedings{zhang2023adaptive,
  title={Adaptive disentangled transformer for sequential recommendation},
  author={Zhang, Yipeng and Wang, Xin and Chen, Hong and Zhu, Wenwu},
  booktitle={Proceedings of the 29th ACM SIGKDD conference on knowledge discovery and data mining},
  pages={3434--3445},
  year={2023}
}

@inproceedings{zhou2023attention,
  title={Attention calibration for transformer-based sequential recommendation},
  author={Zhou, Peilin and Ye, Qichen and Xie, Yueqi and Gao, Jingqi and Wang, Shoujin and Kim, Jae Boum and You, Chenyu and Kim, Sunghun},
  booktitle={Proceedings of the 32nd ACM international conference on information and knowledge management},
  pages={3595--3605},
  year={2023}
}

@inproceedings{buyl2023rankformer,
  title={Rankformer: Listwise learning-to-rank using listwide labels},
  author={Buyl, Maarten and Missault, Paul and Sondag, Pierre-Antoine},
  booktitle={Proceedings of the 29th ACM SIGKDD Conference on Knowledge Discovery and Data Mining},
  pages={3762--3773},
  year={2023}
}

@article{missault2021addressing,
  title={Addressing cold start with dataset transfer in e-commerce learning to rank},
  author={Missault, Paul and de Myttenaere, Arnaud and Radler, Andreas and Sondag, Pierre-Antoine},
  year={2021}
}

@inproceedings{luo2022query,
  title={Query attribute recommendation at amazon search},
  author={Luo, Chen and Headden, William and Avudaiappan, Neela and Jiang, Haoming and Cao, Tianyu and Yin, Qingyu and Gao, Yifan and Li, Zheng and Goutam, Rahul and Zhang, Haiyang and others},
  booktitle={Proceedings of the 16th ACM conference on recommender systems},
  pages={506--508},
  year={2022}
}

@inproceedings{trotman2017architecture,
  title={The Architecture of eBay Search.},
  author={Trotman, Andrew and Degenhardt, Jon and Kallumadi, Surya},
  booktitle={eCOM@ SIGIR},
  pages={88},
  year={2017}
}

@inproceedings{magnani2022semantic,
  title={Semantic retrieval at walmart},
  author={Magnani, Alessandro and Liu, Feng and Chaidaroon, Suthee and Yadav, Sachin and Reddy Suram, Praveen and Puthenputhussery, Ajit and Chen, Sijie and Xie, Min and Kashi, Anirudh and Lee, Tony and others},
  booktitle={Proceedings of the 28th ACM SIGKDD Conference on Knowledge Discovery and Data Mining},
  pages={3495--3503},
  year={2022}
}

@article{ziakis2019important,
  title={Important factors for improving Google search rank},
  author={Ziakis, Christos and Vlachopoulou, Maro and Kyrkoudis, Theodosios and Karagkiozidou, Makrina},
  journal={Future internet},
  volume={11},
  number={2},
  pages={32},
  year={2019},
  publisher={MDPI}
}

@inproceedings{pancha2022pinnerformer,
  title={Pinnerformer: Sequence modeling for user representation at pinterest},
  author={Pancha, Nikil and Zhai, Andrew and Leskovec, Jure and Rosenberg, Charles},
  booktitle={Proceedings of the 28th ACM SIGKDD conference on knowledge discovery and data mining},
  pages={3702--3712},
  year={2022}
}

@article{Zhai2024ActionsSL,
  title={Actions Speak Louder than Words: Trillion-Parameter Sequential Transducers for Generative Recommendations},
  author={Jiaqi Zhai and Lucy Liao and Xing Liu and Yueming Wang and Rui Li and Xuan Cao and Leon Gao and Zhaojie Gong and Fangda Gu and Michael He and Yin-Hua Lu and Yu Shi},
  journal={ArXiv},
  year={2024},
  volume={abs/2402.17152},
  url={https://api.semanticscholar.org/CorpusID:268033327}
}

@article{Xia2023TransActTR,
  title={TransAct: Transformer-based Realtime User Action Model for Recommendation at Pinterest},
  author={Xue Xia and Pong Eksombatchai and Nikil Pancha and Dhruvil Deven Badani and Po-Wei Wang and Neng Gu and Saurabh Vishwas Joshi and Nazanin Farahpour and Zhiyuan Zhang and Andrew Zhai},
  journal={Proceedings of the 29th ACM SIGKDD Conference on Knowledge Discovery and Data Mining},
  year={2023},
  url={https://api.semanticscholar.org/CorpusID:258999754}
}

@inproceedings{zhou2018deep,
  title={Deep interest network for click-through rate prediction},
  author={Zhou, Guorui and Zhu, Xiaoqiang and Song, Chenru and Fan, Ying and Zhu, Han and Ma, Xiao and Yan, Yanghui and Jin, Junqi and Li, Han and Gai, Kun},
  booktitle={Proceedings of the 24th ACM SIGKDD international conference on knowledge discovery \& data mining},
  pages={1059--1068},
  year={2018}
}

@inproceedings{wang2017deep,
  title={Deep \& cross network for ad click predictions},
  author={Wang, Ruoxi and Fu, Bin and Fu, Gang and Wang, Mingliang},
  booktitle={Proceedings of the ADKDD'},
  pages={1--7},
  year={2017}
}



\appendix

\section{Additional Details}
\label{appendix:A}





\subsection{Unified Embedding Table}
We used a unified embedding table with $1.6 \times 10^7$ entries and embedding dimension 1. For each example, we used 6 seeds for the listing ID, 4 seeds for the host ID, and 2 seeds for the s2cell ID. Since we have 14 s2cell levels, this yields $14 \times 2 = 28$ s2cell-related entries retrieved from the unified embedding table.

Based on~\cite{coleman2023unified}, inter-feature collisions can be mitigated by a single-layer neural network because different features are processed by different model parameters. Here, we focus on the intra-feature collision rates for each ID feature given a table of size $N = 1.6 \times 10^7$:

\begin{itemize}[leftmargin=*]
    \item \textbf{Listing ID:} With approximately $M = 10^7$ listings, the load factor is
    \[
        \lambda = \frac{M}{N} = 0.625.
    \]
    The expected collision rate (fraction of entries that collide) under the balls-into-bins approximation is
    \[
        1 - \frac{1}{\lambda}\left(1 - e^{-\lambda}\right) \approx 0.256.
    \]
    With 6 seeds, the probability that all 6 retrieved embeddings collide is
    \[
        0.256^6 \approx 2.9 \times 10^{-4}.
    \]

    \item \textbf{Host ID:} With approximately $M = 5 \times 10^6$ hosts, the load factor is
    \[
        \lambda = \frac{M}{N} = 0.3125.
    \]
    The expected collision rate is
    \[
        1 - \frac{1}{\lambda}\left(1 - e^{-\lambda}\right) \approx 0.144.
    \]
    With 4 seeds, the probability that all 4 retrieved embeddings collide is
    \[
        0.144^4 \approx 4.3 \times 10^{-4}.
    \]

    \item \textbf{s2cell ID:} With approximately $M = 10^6$ s2cells, the load factor is
    \[
        \lambda = \frac{M}{N} = 0.0625.
    \]
    The expected collision rate is
    \[
        1 - \frac{1}{\lambda}\left(1 - e^{-\lambda}\right) \approx 0.030.
    \]
    With 2 seeds, the probability that both retrieved embeddings collide is
    \[
        0.030^2 \approx 9.0 \times 10^{-4}.
    \]
\end{itemize}
  
The overall collision rate is in the $10^{-4}$ level. With multiple seeds, each entity aggregates embeddings from $s$ independently hashed slots, significantly reducing the effective impact of any single collision. 

\begin{figure}[t]
  \centering

  \begin{subfigure}[b]{0.232\textwidth}
    \centering
    \includegraphics[width=0.99\textwidth]{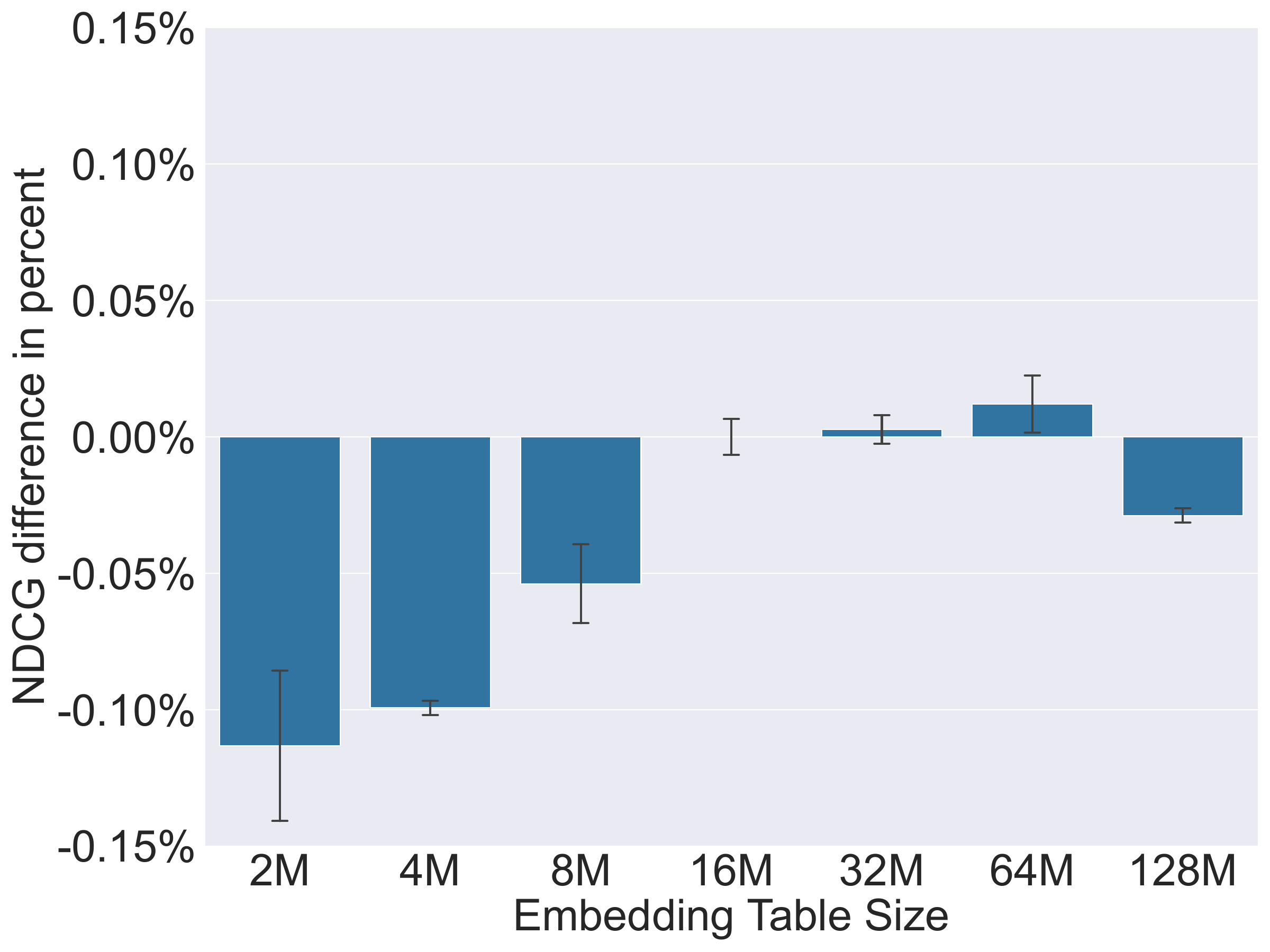}
  \end{subfigure}%
  \begin{subfigure}[b]{0.228\textwidth}
    \centering
    \includegraphics[width=0.99\textwidth]{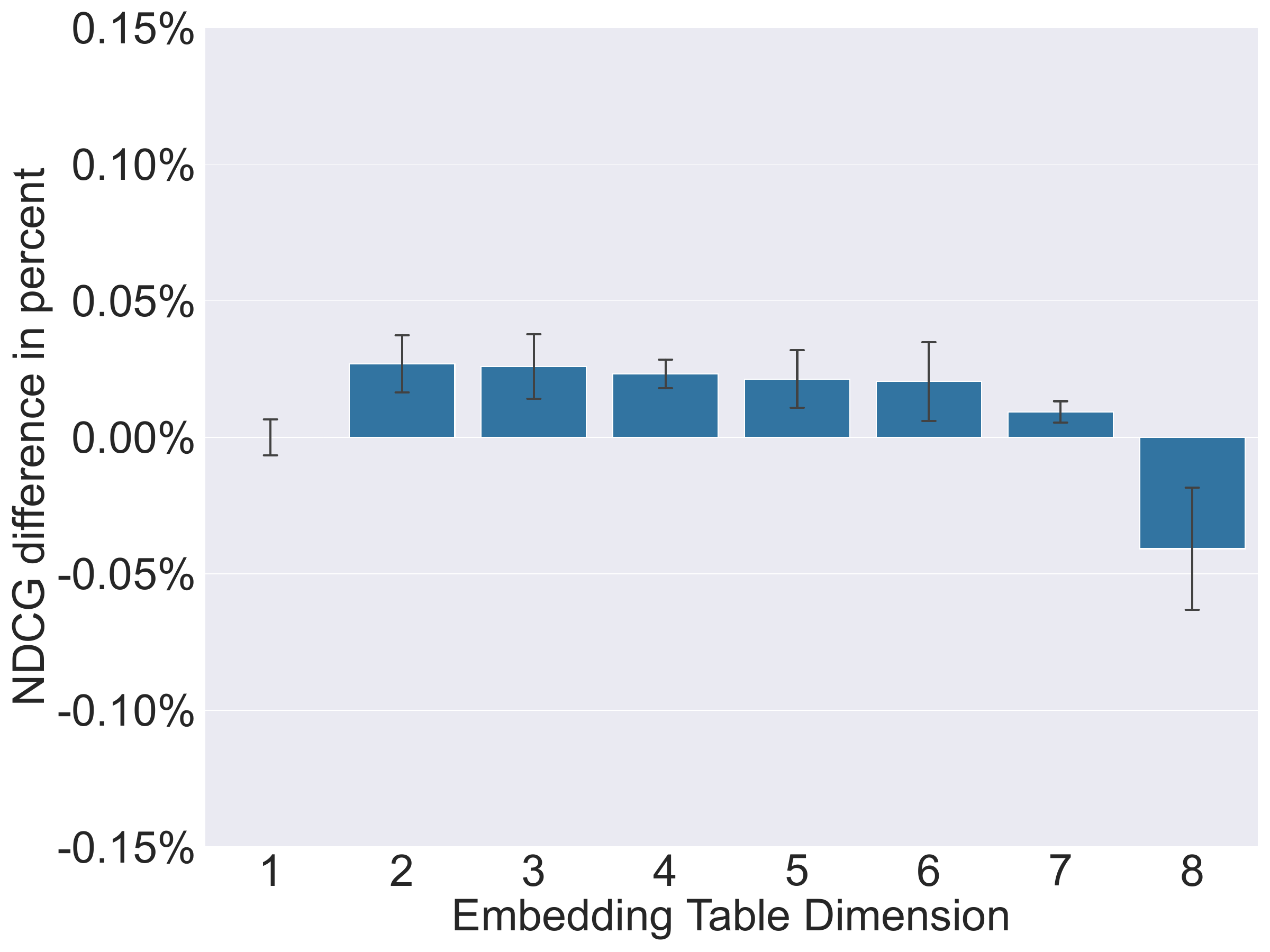}
  \end{subfigure}%
  \caption{NDCG w.r.t. the unified emebdding table size (left) and embedding table dimension (right). We report the percentage difference compared to the overall NDCG. We used a table with dimension 1 and 16M entries for the model tested in production.}
  \label{fig:embedding_table}
\end{figure}

To better understand how collisions affect performance, we plot the impact of the unified embedding table size and embedding dimension in Figure~\ref{fig:embedding_table}. We observe slight NDCG gains as we increase the number of entries or the dimension, but overly large tables lead to NDCG degradation due to overfitting. Since larger tables also increase training and inference cost, we use a table with dimension 1 and 16M entries for the production model as a balance.

\subsection{A/B Testing}

We use an internal A/B testing platform at Airbnb. Traffic is assigned at the guest-ID level using a fixed hashing seed, which deterministically routes each guest to a treatment or control model. Each test usually runs for three weeks.


During each experiment, we monitor a broad set of metrics, including key business metrics, guardrail metrics, and debugging metrics. A model can launch if it improves key business metrics with a p-value below 0.05 and does not violate guardrails. With all these requirements, new ranking model launches are very rare and typically require extensive cross-organization review to ensure the best experience for both guests and hosts.

\section{Additional Discussions \& Results}
\label{appendix:B}

\subsection{Label Attribution Window}

\begin{figure}[t]
  \centering

  \begin{subfigure}[b]{0.232\textwidth}
    \centering
    \includegraphics[width=0.99\textwidth]{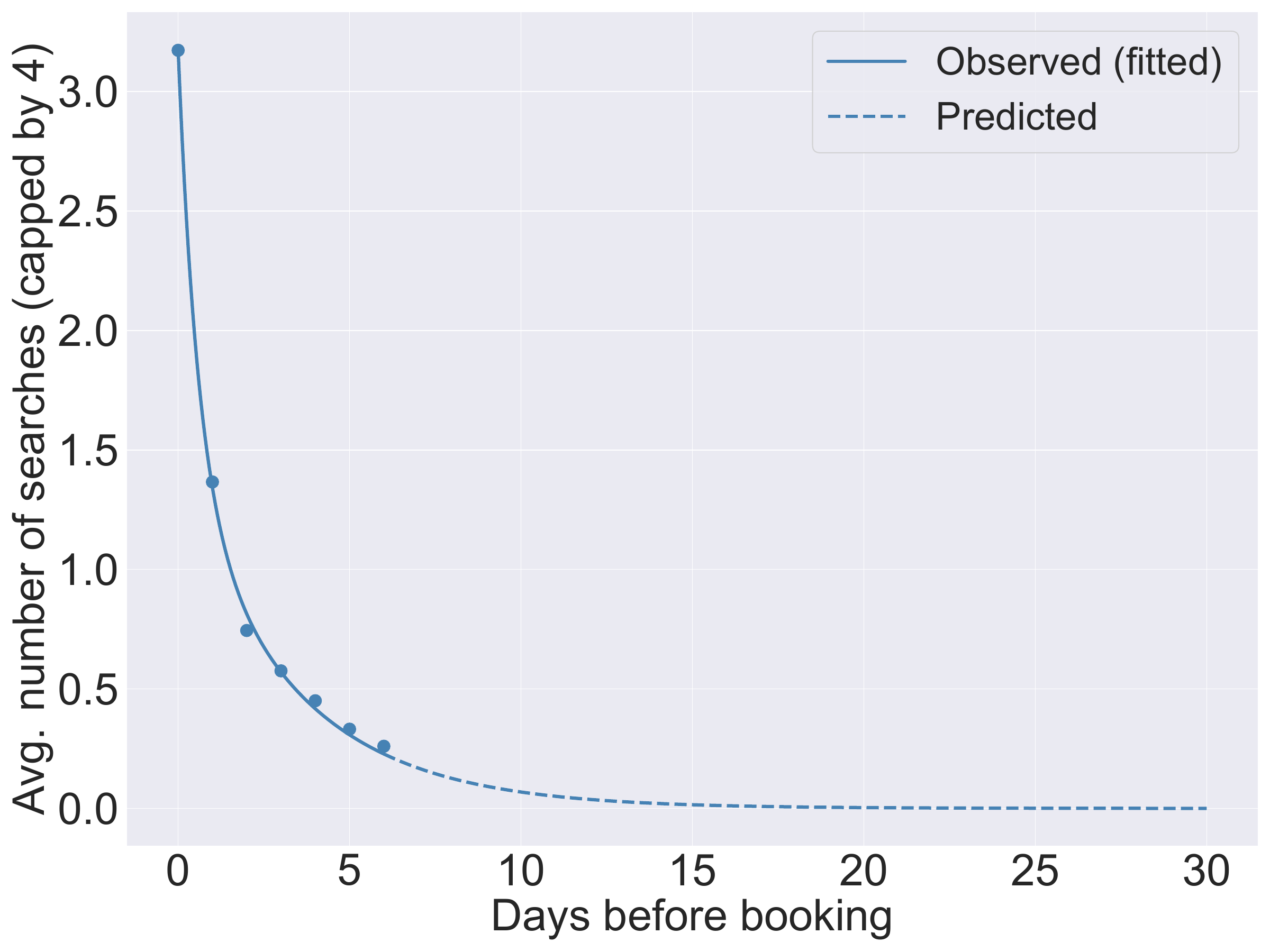}
  \end{subfigure}%
  \begin{subfigure}[b]{0.228\textwidth}
    \centering
    \includegraphics[width=0.99\textwidth]{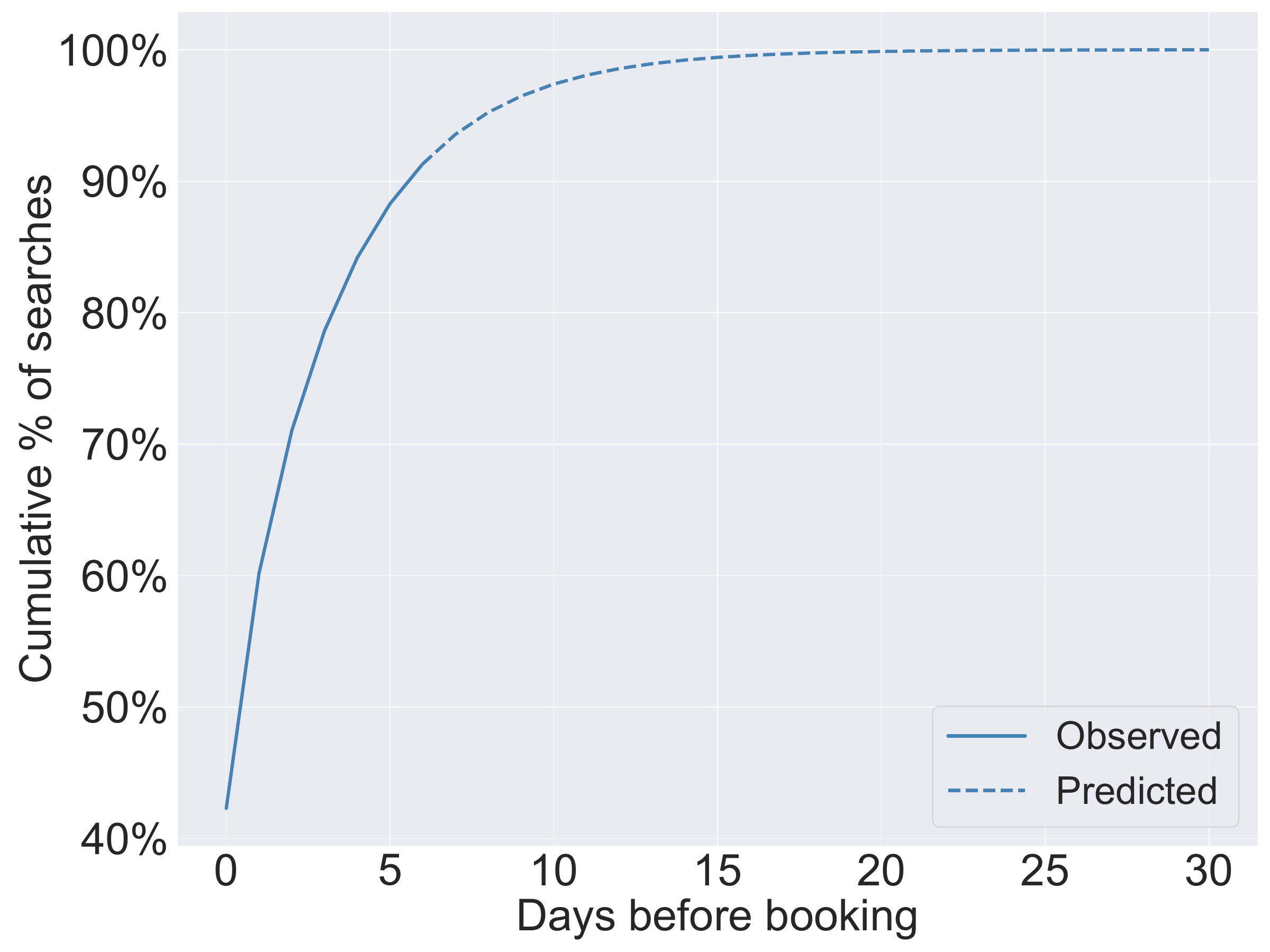}
  \end{subfigure}%
  \caption{The left figure shows the average number of searches (capped at 4 per day) as a function of days before booking. We compute the statistics for days 0–6 directly from our training data and extrapolate for days 6–30 by fitting a power-law distribution. The right figure shows the corresponding cumulative percentage of searches over the 30 days preceding a booking, combining both observed and extrapolated values.}
  \label{fig:label_attribtution}
\end{figure}

Recall that we use a 7-day attribution window. Here, we discuss the trade-offs of using larger or smaller windows. For accuracy, we base this analysis on our training data rather than raw logs, since multiple capping and filtering steps have been applied in our training data. In Figure~\ref{fig:label_attribtution}, we compute the average number of searches (capped at 4 per day) for days 0–6 before a booking using the training data, then fit a power-law distribution to estimate searches for days 6–30. The results show that a 7-day window covers over 90\% of searches, as most guests make booking decisions within a few days.

A larger attribution window may capture more relevant searches and potentially improve performance. However, it also increases data sparsity and significantly lengthens training time, since we batch searches across days. We therefore choose a 7-day window as a practical trade-off between effectiveness and efficiency.

\begin{figure}[t]
  \centering
  \includegraphics[width=0.30\textwidth]{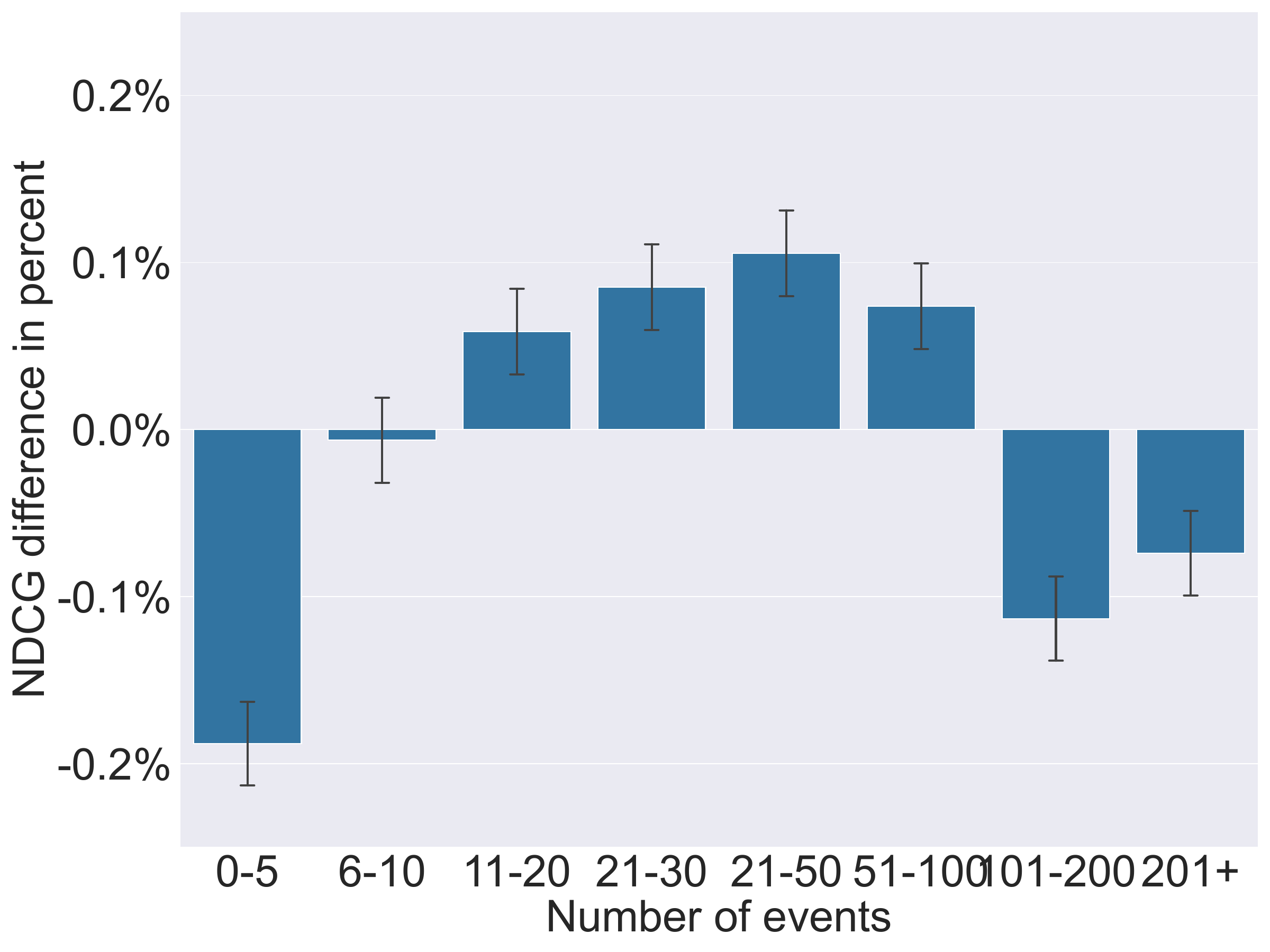}
  \caption{NDCG w.r.t. the total number of events. We report the percentage difference compared to the overall NDCG. To reduce variance and better visualize trends, we bucket sequence lengths, since some lengths are extremely rare. For instance, it is very uncommon for a guest to book a listing after fewer than two prior events (i.e., no past bookings and a decision made after viewing fewer than two listings in the past 21 days). After bucketization, each bucket contains at least 1\% of the data.}
  \label{fig:cold_start}
\end{figure}

\subsection{Cold Start Problem}

Cold-start is a long-standing challenge in recommendation and ranking. Here, we focus on offline NDCG across guest segments with different sequence lengths (Figure~\ref{fig:cold_start}). To reduce variance and better visualize trends, we bucket sequence lengths, since some lengths are extremely rare. For instance, it is very uncommon for a guest to book a listing after fewer than two prior events (i.e., no past bookings and a decision made after viewing fewer than two listings in the past 21 days). After bucketization, each bucket contains at least 1\% of the data.

We observe that offline performance degrades for both very short histories (cold-start) and very long histories (likely due to increased noise in long sequences). A key limitation of this analysis is that NDCG is computed only on sequences that end in a booking, which represent about 5\% of all sequences. It is difficult to assess the impact on the remaining 95\% of non-booking sequences, since we lack labels for them. In future work, we plan to develop better evaluation strategies for cold-start.





\end{document}